\documentclass[a4paper,fleqn]{cas-dc} 

\usepackage[authoryear, round]{natbib}
\usepackage{tabularx, 
            cleveref,
            subcaption,
            import, 
            etoolbox, 
            enumitem}

\usepackage{xcolor}
\usepackage[export]{adjustbox}
            
\definecolor{mydefault}{rgb}{0.0, 0.0, 0.0}

\newcommand \thdr [1] {\textbf {#1}}
\newcommand{\review}[1]{\textcolor{mydefault}{#1}}

\newbool{singlecolumn}
\boolfalse{singlecolumn} 

\Crefname{section}{Section}{Sections}
\crefname{section}{Sect.}{sections}
\Crefname{table}{Table}{Tables}
\crefname{table}{Tbl.}{tables}
\Crefname{figure}{Figure}{Figures}
\crefname{figure}{Fig.}{figures}
\Crefname{equation}{Equation}{Equations}
\crefname{equation}{Eq.}{equations}

\begin{document}

\let\WriteBookmarks\relax
\def\floatpagepagefraction{1}
\def\textpagefraction{.001}

\shorttitle{SeagrassFinder}    
\shortauthors{Elsäßer et al.}  


\title [mode = title]{SeagrassFinder: Deep Learning for Eelgrass Detection and Coverage Estimation in the Wild}  


\author[1]{Jannik Elsäßer}[orcid=0009-0002-2193-4147]
    \fnmark[1]
    \ead{jael@dhigroup.com}
    \credit{Writing – original draft, Software, Data curation, Methodology, Conceptualization, Visualization}
    \affiliation[1]{organization={DHI A/S},
            addressline={Agern Alle 5}, 
            city={Hørsholm },
            postcode={2970}, 
            country={Denmark}}

\author[2]{Laura Weihl}[orcid=0000-0002-8916-3468]
    \ead{lawe@itu.dk}
    \cormark[1]
    \credit{Writing – review \& editing, Methodology, Supervision, Validation, Visualization}
    \affiliation[2]{organization={Computer Science Department, IT University of Copenhagen},
                addressline={Rued Langaards Vej 7}, 
                city={Copenhagen},
                postcode={2300}, 
                country={Denmark}}
    
\author[2]{Veronika Cheplygina}[orcid=0000-0003-0176-9324]
    \ead{vech@itu.dk}
    \credit{Writing – review \& editing, Methodology, Supervision, Conceptualization}
    
\author[1]{Lisbeth Tangaa Nielsen}
    \ead{litn@dhigroup.com}
    \credit{Writing – review \& editing, Methodology, Validation}



\begin{abstract}
Seagrass meadows play a crucial role in marine ecosystems, providing benefits such as carbon sequestration, water quality improvement, and habitat provision. Monitoring the distribution and abundance of seagrass is essential for environmental impact assessments and conservation efforts. However, the current manual methods of analyzing underwater video data to assess seagrass coverage are time-consuming and subjective. This work explores the use of deep learning models to automate the process of seagrass detection and coverage estimation from underwater video data. We create a new dataset of over 8,300 annotated underwater images, and subsequently evaluate several deep learning architectures, including ResNet, InceptionNetV3, DenseNet, and Vision Transformer for the task of binary classification on the presence and absence of seagrass by transfer learning. The results demonstrate that deep learning models, particularly Vision Transformers, can achieve high performance in predicting eelgrass presence, with AUROC scores exceeding 0.95 on the final test dataset. The application of underwater image enhancement further improved the models' prediction capabilities. Furthermore, we introduce a novel approach for estimating seagrass coverage from video data, showing promising preliminary results that align with expert manual labels, and indicating  potential for consistent and scalable monitoring. The proposed methodology allows for the efficient processing of large volumes of video data, enabling the acquisition of much more detailed information on seagrass distributions in comparison to current manual methods. This information is crucial for environmental impact assessments and monitoring programs, as seagrasses are important indicators of coastal ecosystem health. This project demonstrates the value that deep learning can bring to the field of marine ecology and environmental monitoring.
\end{abstract}


\begin{graphicalabstract}
    \includegraphics{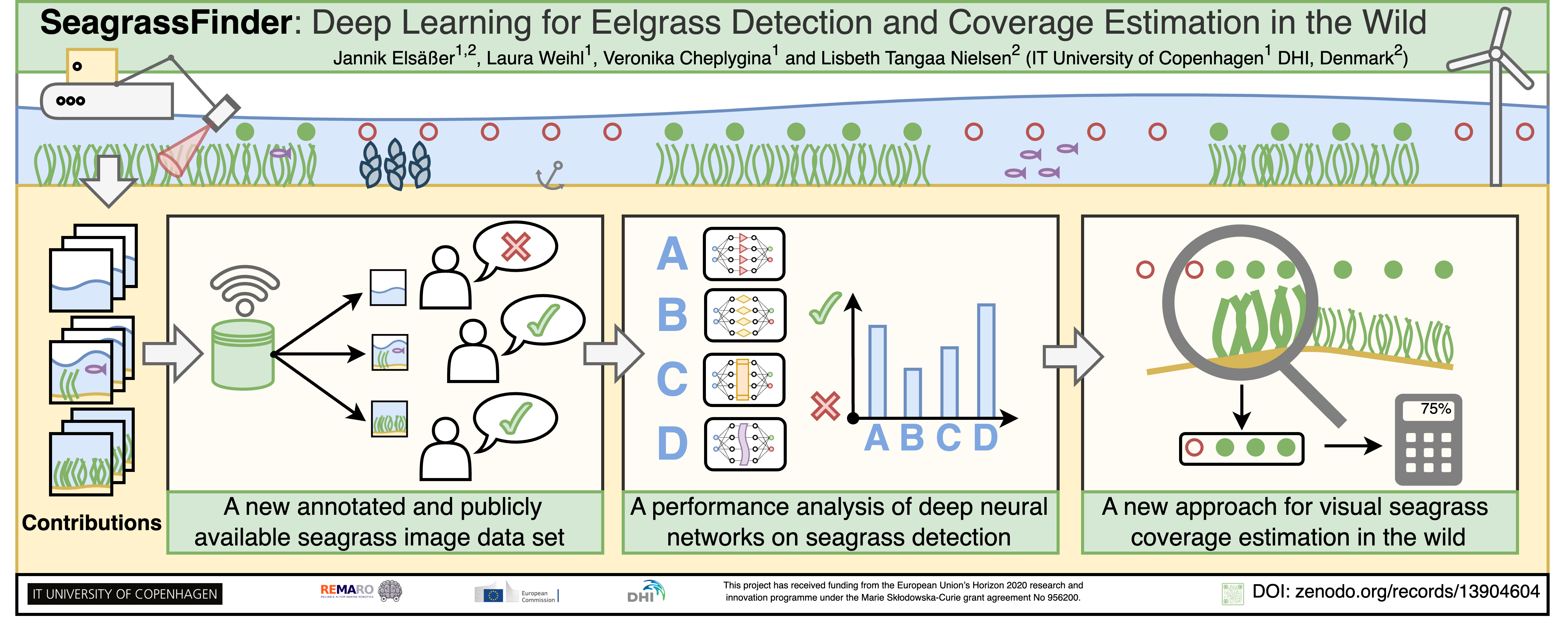}
\end{graphicalabstract}


\begin{highlights}
    \item Deep neural networks can effectively automate seagrass presence detection in underwater videos from environmental impact assessments, reducing reliance on manual annotations.
    \item Transfer learning combined with underwater image enhancement techniques improves deep learning capabilities for seagrass detection tasks. 
    \item Our custom data annotation platforms support labeling of large amounts of data collected from environmental impact assessments.
    \item Our methodology serves as a supportive tool for marine biologists, enabling more consistent, scalable, and objective environmental monitoring.
\end{highlights}



\date{\today}

\maketitle


\section{Introduction}
\label{sec:intro}

\review{Seagrasses are a type of submerged aquatic vegetation (SAV) found in shallow waters alongside every continent in the world except Antarctica \citep{novak2020submerged}. As a marine flowering plant, they are of fundamental importance to the health and integrity of coastal ocean habitats, biodiversity, global climate, and human economic stability as well as global food security \citep{ unsworth2019global, cullen2014seagrass}. Seagrass meadows are monitored worldwide as part of environmental impact assessments (EIAs) and environmental monitoring programs because~they provide important ecosystem services, such as carbon~sequestration, water quality improvement, providing food and habitat, and acting as a biological indicator of local ecosystems \citep{novak2020submerged}. Seagrass meadows serve as vital hubs of biodiversity, offering nourishment and shelter to a wide range of organisms, including endangered species like dugongs and commercially valuable ones such as fish and shrimp, as well as microbes and invertebrates \citep{fry1979animal}. They are estimated to support up to 20\% of global fishery productivity \citep{unsworth2019seagrass}, making them key contributors to marine food webs and fisheries management. Although seagrasses cover only 2\% of the ocean's seabed, they account for over 15\% of the carbon annually sequestered in marine environments \citep{fourqurean2012seagrass}. Despite ongoing conservation efforts, seagrass meadows are declining globally at an alarming rate due to human activities and coastal development, which is further exacerbated by the effects of climate change \citep{waycott2009accelerating}. These factors collectively contribute to a stark decline in water quality, and threaten the long-term survival of seagrass habitats. Due to seagrasses immediate vulnerability to environmental factors and their key ecosystem functions, they are important biological indicators and ecological status of coastal conditions \citep{orth2006global,balsby2013sources}.}
\looseness -1

\review{Estimating the abundance of seagrass meadows is critical to judging seagrasses' ecosystem functions \citep{hemminga2000seagrass}. The fundamental challenge here is to acquire suitable data in a structured, automated and efficient manner \citep{unsworth2019global}. Methods for seagrass monitoring include analysis of images from remote sensing \citep{veettil2020opportunities}, for example collected from satellite \citep{riegl2005detection}, and from aerial footage \citep{roelfsema2015field}. Yet these methods fail to distinguish between different types of seagrass and lack fine-grained resolution. Additionally, aerial surveys are heavily limited by water turbidity and favorable weather conditions. Traditionally, scuba divers have been deployed to conduct seagrass monitoring \citep{sengupta2020seagrassdetect}, however this is extremely restricted due to the required amount of human labor.  There is a trend towards monitoring with autonomous underwater vehicles which adds new challenges and possibilities to automated seabed mapping \citep{moniruzzaman2019imaging}}.
\looseness -1

\review{Environmental Impact Assessments (EIAs) conducted offshore are intended to monitor and manage the environmental impacts of coastal developments such as offshore construction sites in order to ensure environmental compliance with different stakeholders across the marine infrastructure sector. During EIAs, monitoring is often conducted in the form of video transects, among other methods, to determine the seagrass abundance over a range of water depths \citep{short2001global}. To achieve this, a camera is mounted onto a sled and towed behind a vessel to capture underwater video footage. The recorded videos are then annotated on the presence or absence of seagrass by marine biologists. This method provides several advantages; a large spatial coverage,  more fine-grained image data of the underwater environment, and the possibility of repeated monitoring in a structured and efficient manner. Similarly to other image-based monitoring methods, the subsequent data labeling task is extremely tedious as it is done manually and thus requires a vast amount of working hours.}
\looseness -1

\begin{figure*}[!ht]
    \centering
    \includegraphics[width=\textwidth]{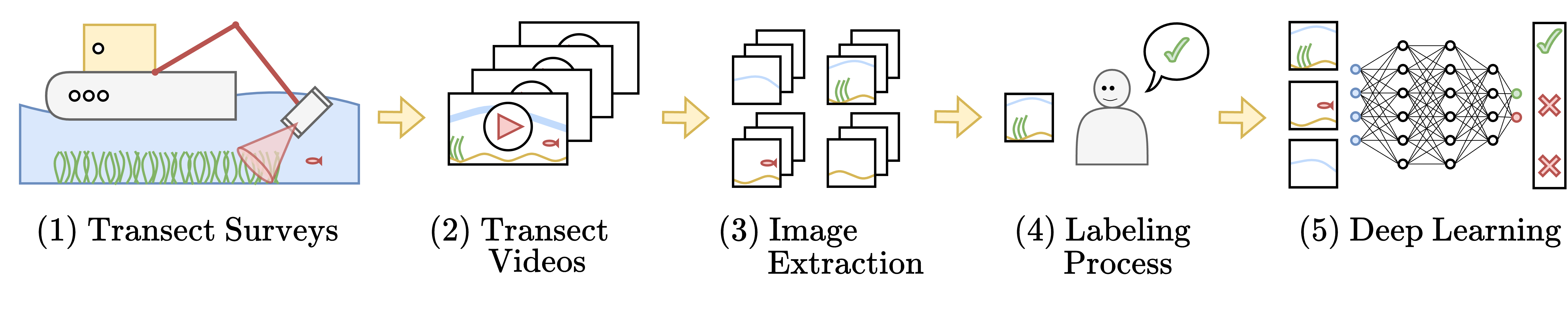}
    \caption{The SeagrassFinder Project Pipeline: (1) a vessel performs transects surveys by towing a sled with a camera along the seabed and (2) records videos of the underwater environment. The extracted images (3) are then labeled by human annotators (4) and used to train deep neural networks (DNNs) on the task of detecting eelgrass (5). A trained DNN can now be used to replace the time-consuming manual annotation process.}
    \label{fig:sketch-pipeline}
\end{figure*}

\review{With the advent of deep learning, our capacity to analyze and process large amounts of data has been transformed significantly. Specifically deep neural networks (DNNs) have shown remarkable success in a wide range of computer vision tasks such as image classification, semantic segmentation and object detection. DNNs achieve this by learning information directly from raw pixel input data without the need for manual feature engineering. Especially convolutional neural networks \citep{krizhevsky2012imagenet} have massively improved our ability to automatically extract relevant features and have become the new standard for many computer vision tasks. DNN training and prediction for the task of seagrass monitoring is a promising direction, because DNNs can conduct image classifications in an automated, repeatable and efficient manner and offer generalizability across a variety of image qualities suffered from underwater images, such as blur, color-distortion or low contrast and low light. Meanwhile, seagrass stalks can have complex structures and shapes, thus, even single-species seagrass detection is a hard challenge. Moniruzzaman et al. \citep{moniruzzaman2019faster} detect single seagrass stalks of Halophila ovalis with a Faster R-CNN model on their own compiled data set ECUHO, containing images in a laboratory setting and images taken in the wild. Noman et al. \citep{noman2023improving} achieve a higher mAP performance of 0.484 on the ECUHO data set, with two popular object detection models, YOLOv5 and EfficientNetD7. Sengupta et al. \citep{sengupta2020seagrassdetect} also investigate single-species detection. They use traditional feature extraction methods in combination with a Gaussian mixture model for binary classification on absence or presence of seagrass.}
\looseness -1

\review{Multi-species seagrass classification is often evaluated on the patch-based DeepSeagrass data set \citep{raine2020multi}, on which the authors achieve an accuracy of 92.4\% via semi-automatically patch labeling and a convolutional DNN. In the work by Noman et al. \citep{noman2021multi}, the authors achieve overall higher accuracy with a semi-supervised learning approach. Latest advances by unsupervised curriculum learning by Abid et al. \citep{abid2024seagrass} perform best at 90.12\% precision.}
\looseness -1

\review{Prior work on image segmentation includes feature extraction in combination with logistic model trees \citep{bonin2016towards}, superpixel-based solutions \citep{reus2018looking}, later revised by Weidmann et al. \citep{weidmann2019closer},  and patch-based approaches \citep{burguera2020segmentation}, reaching 95.5\% of accuracy. Further work on a semantic segmentation on a pixel basis was conducted first with a VGG16 encoder and FCN8 decoder \citep{martin2018deep}, later with an attention-based encoder-decoder modules \citep{wang2020compact}, as well as with an encoder similar to MobileNet \citep{wang2020real}. Most recently, Raine et al. \citep{raine2024image} have combined large language models as a supervisory signal with unsupervised contrastive pre-training for coarse seagrass segmentation.}
\looseness -1

We address the challenge of estimating seagrass presence and coverage in the wild from vast amounts of raw, unstructured and unfiltered video data taken in realistic environmental conditions. We achieve this by deploying a custom-made annotation platform for efficient manual labeling, as well as training state-of-the-art deep learning models for classifying whether eelgrass is present or absent in an image. To the best of our knowledge we are the first to deploy Vision Transformers \citep{dosovitskiy2020image} to detect the presence of eelgrass in underwater images. \review{In this work, we emphasize the distinction between spatial seagrass coverage - referring to the actual vegetation density - and pixel-based coverage, which determines the proportion of image pixels classified as seagrass. This is often overlooked in current contributions, yet we argue spatial seagrass coverage holds greater ecological relevance, as it reflects actual vegetation density rather than just pixel-level detection.} Specifically we contribute the following:
\looseness -1

\begin{itemize}[leftmargin=*]
    \item a strategy for effectively streamlining human labeling efforts of human experts and non-experts on underwater seagrass images and compiling the annotated data set into an efficient deep learning data pipeline (see \cref{fig:sketch-pipeline}), 
    \item a novel annotated image data set of eelgrass under realistic non-artificial lighting conditions extracted from raw videos of underwater transect surveys in the wild\footnote{\href{zenodo.org/records/13904604}{DOI: zenodo.org/records/13904604}},
    \item an analysis of state-of-the-art DNN models and their ability to classify a range of challenging underwater images according to the presence of eelgrass using a transfer learning mechanism, as well as an investigation of the effect of UW image enhancement on DNN performance,
    \item a new experimental approach for further utilizing the above-mentioned DNN models for estimating visual eelgrass coverage for an image stream from binary predictions as a post-processing step,
    \item a pipeline which dramatically reduces the time and financial resources required for human annotation, and paves the way for automated environmental monitoring for example by means of autonomous underwater vehicles.
\end{itemize}

The remainder of this paper is structured as follows: \cref{sec:mat-and-met} describes the data strategy, our deep learning methodology for eelgrass detection, as well as our novel eelgrass coverage estimation. We present our experimental design in \cref{sec:experiments} and our results in section  \cref{sec:results}. Finally, we present our discussion and conclusion in \cref{sec:discussion} and \cref{sec:conclusion}.
\looseness -1

\label{sec:conclusion}


\section{Materials and Methods}
\label{sec:mat-and-met}

In this section we present our data strategy, followed by our deep learning methodology on the task of eelgrass detection and finally our novel eelgrass coverage estimation as a postprocessing step.


\subsection{Data Acquisition}
\label{subsec:data-acquisition}

\begin{figure}
        \centering
    \begin{subfigure}[b]{\ifbool{singlecolumn}{0.35\textwidth}{0.9\columnwidth}}
        \centering
        \begin{subfigure}{\linewidth}
            \includegraphics[width=\linewidth]{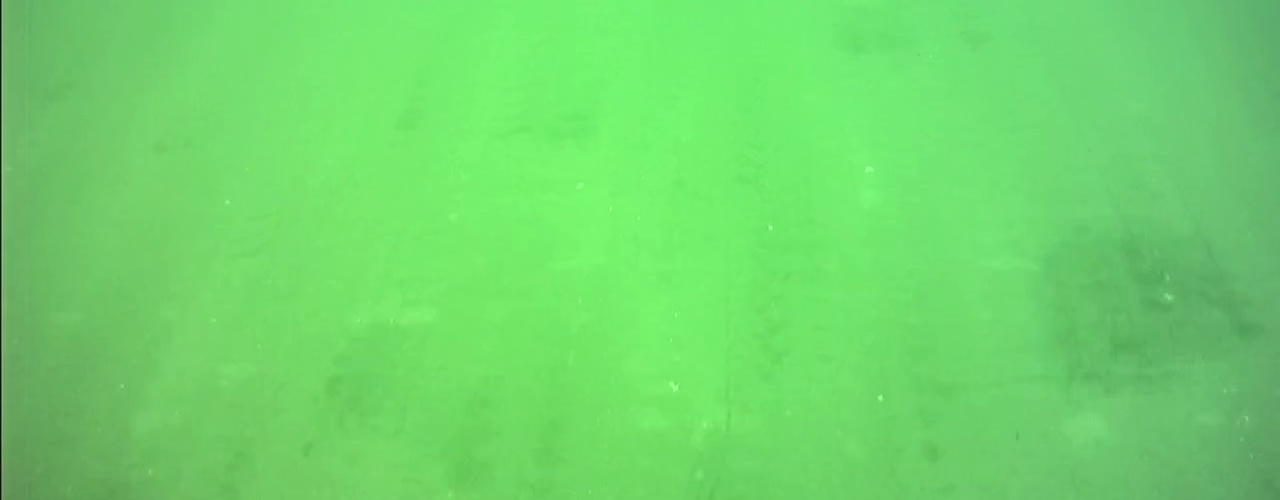}
            \caption{Unvegetated}
            \label{fig:no-sav}
        \end{subfigure}

        \begin{subfigure}{\linewidth}
            \includegraphics[width=\linewidth]{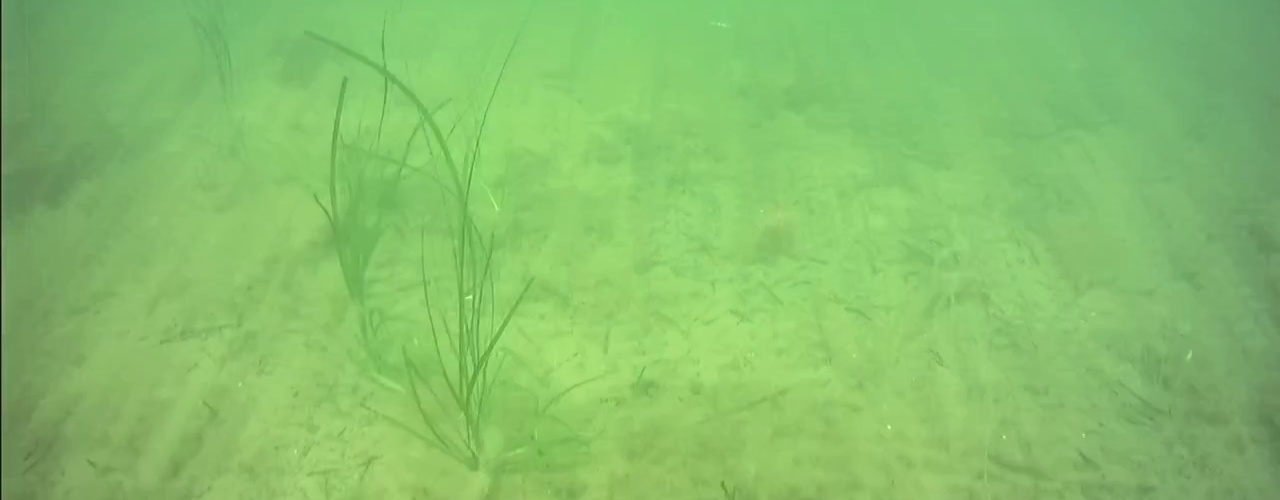}
            \caption{Sparse SAV}
            \label{fig:sparse-sav}
        \end{subfigure}

        \begin{subfigure}{\linewidth}
            \includegraphics[width=\linewidth]{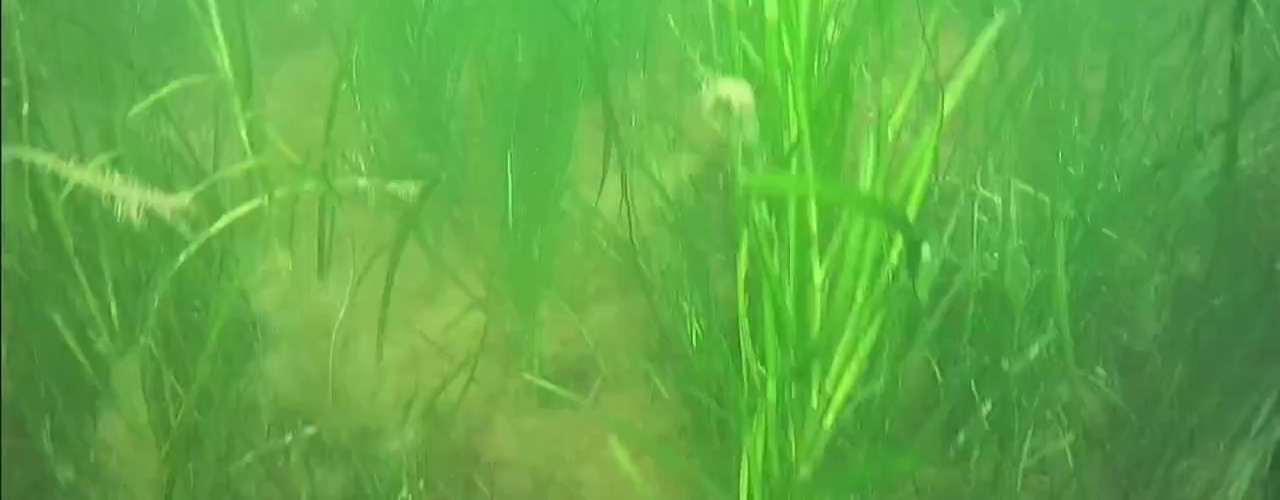}
            \caption{Dense SAV}
            \label{fig:dense-sav}
        \end{subfigure}
    \end{subfigure}%
    \ifbool{singlecolumn}{\hfill}{}
    \begin{subfigure}[b]{\ifbool{singlecolumn}{0.638\textwidth}{0.9\columnwidth}}
        \centering
        \includegraphics[width=\linewidth, clip, trim={0 2mm 0 2mm}]{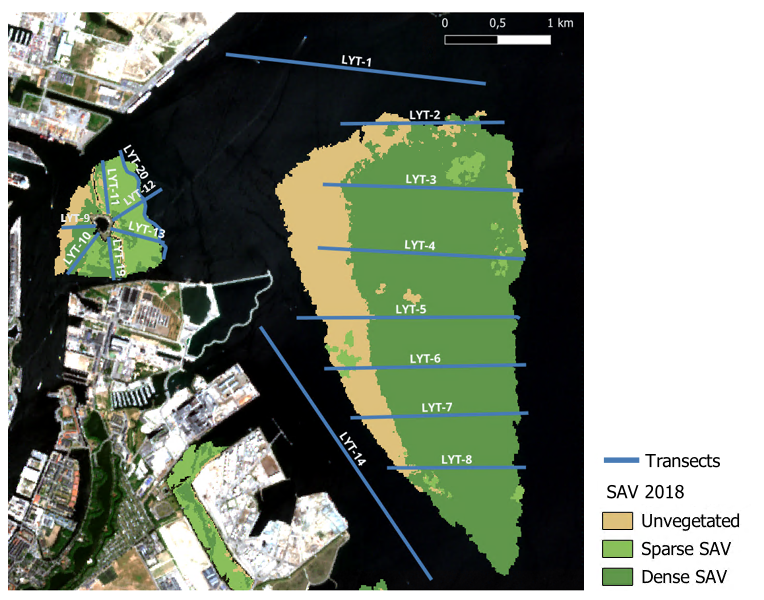}
        \caption{Transect Survey Map}
        \label{fig:transect-map}
    \end{subfigure}
        
    \caption{(a) - (c) Sample images of areas with no, sparse and dense submerged aquatic vegetation (SAV). (d) A map of the Copenhagen harbor with transect survey lines (blue) and SAV data overlaid. All SAV maps are based on \textcopyright \,  Copernicus Sentinel-2 data from 2023-06-13. All SAV maps are based on analysis of Sentinel-2 imagery from 2018 (map created with QGIS v.3.30.2).}
    \label{fig:sav-mapping}
\end{figure}

The dataset used in this project originates from DHI's Environmental Impact Assessment (EIA) and baseline reporting of the Lynetteholm project, a new artificial peninsula being created to extend the coastline of northern Copenhagen in Denmark. We focus on one species of seagrass, namely \textit{Zostera marina}, commonly known as ``eelgrass'' which is the target specific seagrass species monitored in the Lynetteholm EIA (see samples in \cref{fig:no-sav,fig:sparse-sav,fig:dense-sav}. We compile a novel annotated image dataset of \textit{Zostera marina} from six transect surveys in- and outside the Copenhagen harbor, covering an area of roughly $10\textrm{km}^2$. In total we consider 20 transect surveys (marked blue in \cref{fig:transect-map}) over two consecutive days in June 2023 during daytime. Each transect survey produces a transect video of the local benthic environment, the lowest level in a body of water, close to the seabed. During a transect survey, an underwater camera is mounted on a sled, a mechanical device, designed to hover at a roughly consistent height above the seabed while towed behind a survey vessel along a planned route of interest. The sled ensures the camera maintains a consistent angle and prevents it from swinging in the water current. A technician or biologist on the vessel monitors the depth of the sled and monitors its height above the seafloor, adjusting its height to avoid obstacles or to perform necessary maintenance such as cleaning the camera lens. After the field data collection, a marine biologist reviews the camera footage and adds annotations on seagrass abundance to a spreadsheet. These annotations include timestamps, the vessel's location, and any notable changes in eelgrass coverage or other observations. All videos are filmed with an underwater camera with a 1/3” 2 Mega pixel Sony CMOS sensor.

As part of the EIA of the Lynetteholm project, DHI performs 20 transect surveys annually, each producing one transect video. Out of all 20 transect videos of 2023, we choose a subset of six transect videos, based on two factors. We are primarily interested in transect videos where the visual content shows a high degree in variability. We speculate this enhances a DNN's capability for generalization and classification accuracy, for example when confronted with changes in seabed textures. We additionally choose transect videos for their roughly equal class distribution of eelgrass present vs. absent images, in order to avoid model bias in the subsequent DNN training. A balanced data set ensures that a DNN learns both classes well, without favoring a more frequently represented class. We estimate this based on earth observation data provided by DHI as part of the EIA, which showed Submerged Aquatic Vegetation (SAV) areas of interest (see \cref{fig:transect-map}), and initial data inspection. Each selected transect video is between 5-60 minutes long and recorded at 30 FPS while the survey vessel sails at roughly 1m/s. We extract frames at 5 FPS from the survey videos in order to strike a balance between the volume of data and the informational content. This ensures that consecutive frames are sufficiently distinct. In total we extracted 340,000 images from our selected transect videos. We crop all frames to 1280x500 pixels to remove any overlay information (see examples in \cref{fig:sav-mapping}). From each transect video we randomly sample images to create the dataset, resulting in a total of 16,000 images.
\looseness -1

\begin{table}[t!]
    \centering
    \caption{Transect Details. All transects were filmed in June 2023 over two consecutive days during daytime.}
    \begin{tabularx}{\columnwidth}{p{10.5mm}| p{12mm} X X p{6mm}}

        \thdr{Transect Name}
        & \thdr{Length (km)}
        &\thdr{Min. Water Depth (m)}
        &\thdr{Max. Water Depth (m)}
        & \thdr{Video (min)} \\\hline
    
        \thdr{LYT-5} & 2.00 & 5.5 & 12.4 & 20 \\
        \thdr{LYT-9} & 0.29 & 2.7 & 7.4 & 5 \\
        \thdr{LYT-10} & 0.45 & 3.1 & 6.0 & 10 \\
        \thdr{LYT-12} & 0.50 & 3.4  & 12.8 & 10 \\
        \thdr{LYT-14} & 3.00 & 13.8  & 14.1  & 60 \\
        \thdr{LYT-20} & 1.20 & 9.5  & 9.6 & 45 \\

  \end{tabularx}

  \label{tab:dataset}

\end{table}


\subsection{Data Annotation and Platform}
\label{sec:data-annotation}

With our data annotation platform SeagrassFinder, our goal is to create a streamlined annotation process to enable easy access via a browser on any device of an annotator's preferred choice. We considered using already existing annotation platforms however found that they did not provide a UI which was both easy to access and mobile friendly, and therefore developed our own. We named this app SeagrassFinder, to keep the URL of the app simple and memorable. In the context of the paper, we refer to the annotation platform as SeagrassFinder AP. SeagrassFinder AP is a Python Streamlit application deployed as an Azure App Service, connected with an Azure SQL Database for saving the annotations, with individual images hosted on an Azure Storage account. The aim is to create an intuitive and self-explanatory interface that enables a fast labeling process. Since we expect a relatively small count of distinct users we decide to only require users to enter their name without a user having to sign up for an account. \looseness -1

\begin{figure}[!ht]
    \centering
    \includegraphics[cfbox=black 1pt 1pt, width=\ifbool{singlecolumn}{0.4\textwidth}{\columnwidth}]{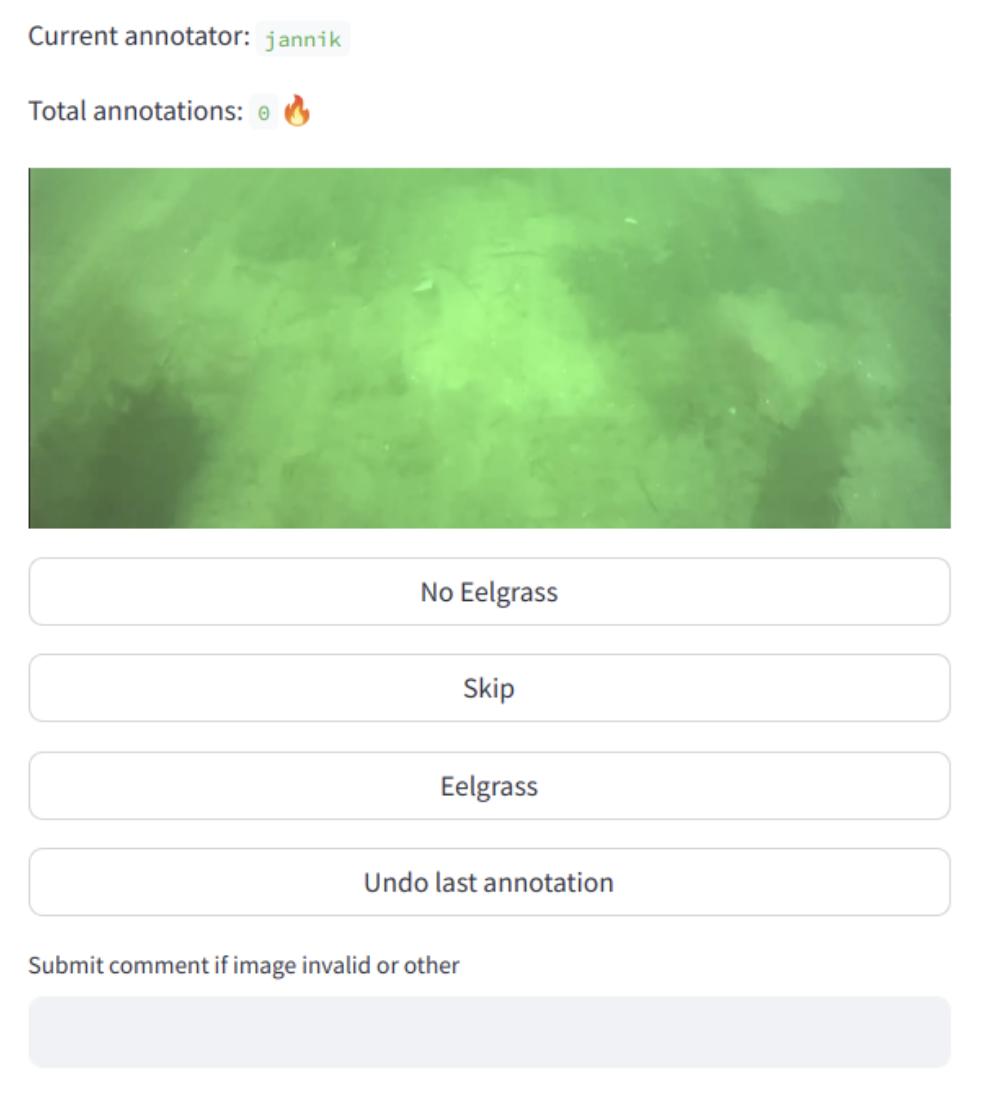}
    \caption{A screenshot of the SeagrassFinder Annotation Platform accessed through a browser window. The labeling interface is simple and intuitive.}
    \label{fig:seagrassfinder_annot}
\end{figure}

The interface of SeagrassFinder AP is designed to be straight-forward with a minimal amount of text to keep the barrier of entry for a new user as low as possible (see \cref{fig:seagrassfinder_annot}). We also keep the explanation extremely brief and simplify the task instructions by adding visual guidance with concrete image examples, both with eelgrass present and absent. Alongside the two label buttons we add a button to skip the current sampled image as well as a button to undo the previous annotation. A user can also submit a comment if an image is considered invalid. For each annotation task we uniformly sample an image from the data set and display it to the annotator to maximize objectivity and to avoid showing consecutive frames which could potentially add labeling bias. To motivate users to annotate as many images as possible we add a leaderboard to the platform, to spark competitiveness between users which proved highly effective.
\looseness -1

\subsection{Dataset Compilation and Training Strategy}

A detailed overview of our data set strategy for the DNN training and testing pipeline can be seen in \cref{tab:dataset-overview}. First we manually remove all images taken above sea-level manually to mitigate inclusion of non-relevant features. Additionally we filter out all images with disagreeing annotations. The final dataset consists of 8324 images, 52\% of which are labeled as ``Eelgrass Present''. To avoid data leakage between train and test sets we do not mix images between different transects. Throughout the project, we hold out transect LYT-9 as the final test dataset. In all experiments we split the training data into validation sets of 20\% using random splits.
\looseness -1

\begin{table*}[t!]
  \caption{SeagrassFinder data strategy: each transect results in a video from which we extract and annotate a subset of images. We list the percentage of images with eelgrass labels, the datasets used for initial hyperparameter tuning and cross validation (cross val.). We excluded LYT-5 and LYT-14 from cross validation because of their high class imbalance. LYT-9 was chosen as the final test set and therefore excluded from all other training/testing procedures. All remaining images were used in the final model training.}

  \begin{tabularx}{\linewidth}{
    >{\small}p{13mm}| 
    >{\small\centering\arraybackslash}p{13mm}
    >{\small\centering}p{16mm}
    >{\small\raggedleft}p{13mm}
    >{\small\centering\arraybackslash}X
    >{\small\centering\arraybackslash}X
    >{\small\centering\arraybackslash}X
    >{\small\centering\arraybackslash}X
    >{\small\centering\arraybackslash}X
  }
    \thdr{Transect Name}
    & \thdr{\#Images} & \thdr{\#Images Annotated} & \thdr{Eelgrass Present} & \thdr{Initial Train Set} & \thdr{Initial Test Set} & \thdr{Cross Val. Set} & \thdr{Final Train Set} & \thdr{Final Test Set} \\\hline \\[-2mm]
    
    \thdr{LYT-5} & 4000 & 2116  & 93.5\% & \checkmark & & & \checkmark & \\
    \thdr{LYT-9} & 2000 & 1044 & 53.5\% & & & & & \checkmark  \\
    \thdr{LYT-10} & 2000 & 1066 & 61.5\% & \checkmark & & \checkmark & \checkmark &\\
    \thdr{LYT-12} & 2000 & 1069 & 52.2\%  & \checkmark & & \checkmark & \checkmark &\\
    \thdr{LYT-14} & 4000 & 2126 & 2.6\% & \checkmark & & & \checkmark &  \\
    \thdr{LYT-20} & 2000 & 1079 & 60.2\% & & \checkmark & \checkmark & \checkmark &  \\\hline \\[-2mm]

    \thdr{Total} & 16000 & 8500 & 53.9\% & 6377 & 1079 & 3214 & 7456 & 1079 \\
    
  \end{tabularx}
  \label{tab:dataset-overview}

\end{table*}


\review{\subsection{Data Preprocessing}}
\label{sec:data-preprocessing}

\begin{figure}[!ht]
  \centering
    \includegraphics[width=\ifbool{singlecolumn}{0.7\textwidth}{\columnwidth}]{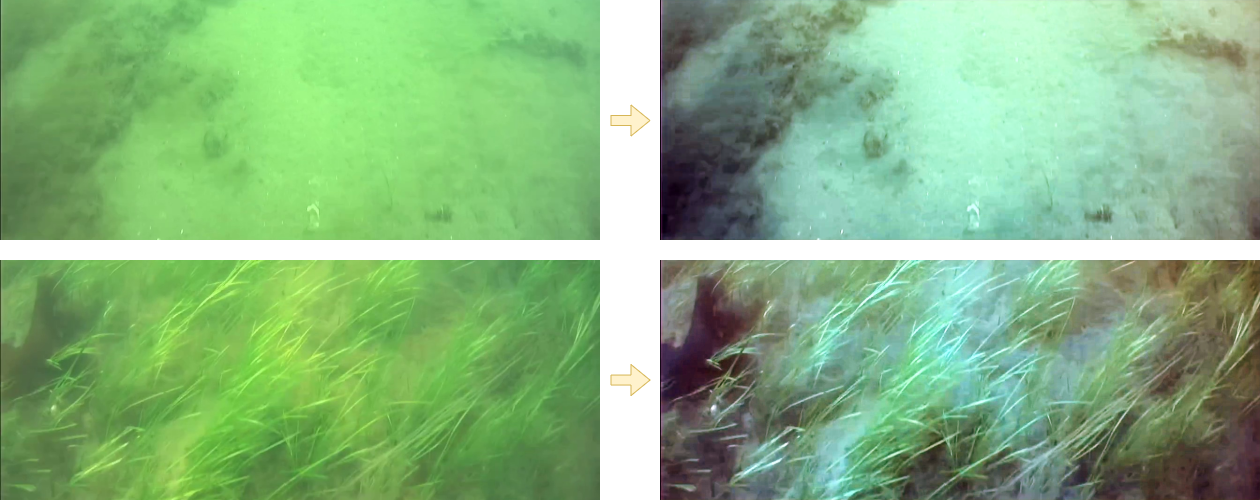}
  \caption{Two images before (left) and after (right) applying underwater image enhancement.}
  \label{fig:eelgrass-enhanced}
\end{figure}

\review{One fundamental challenge in assessing underwater image data is the light attenuation in water with increasing water depth. To mitigate these factors and to increase visual quality of the images, we employ a DNN-based underwater (UW) image enhancement tool called DeepWave-Net \citep{sharma2023wavelength}. We are curious whether this could have a positive effect on prediction accuracy of our DNN models. We conduct all model training and testing procedures with and without the application of UW-enhancement (see \Cref{fig:eelgrass-enhanced}). The UW-enhanced images show a wider range of colors, a better contrast between the green and red channels on the RGB color range. This enables easier differentiation between the green/brown filamentous algae from the eelgrass itself, as well as the sugar kelp within the image. We notice that the effect is even more noticeable, when applying the model on images at a greater water depth.}
\looseness -1

\subsection{Deep Learning for Eelgrass Detection}
\label{sec:deep-learn}

\begin{figure}[!ht]
    \centering
    \includegraphics[width=\ifbool{singlecolumn}{0.6\textwidth}{\columnwidth}]{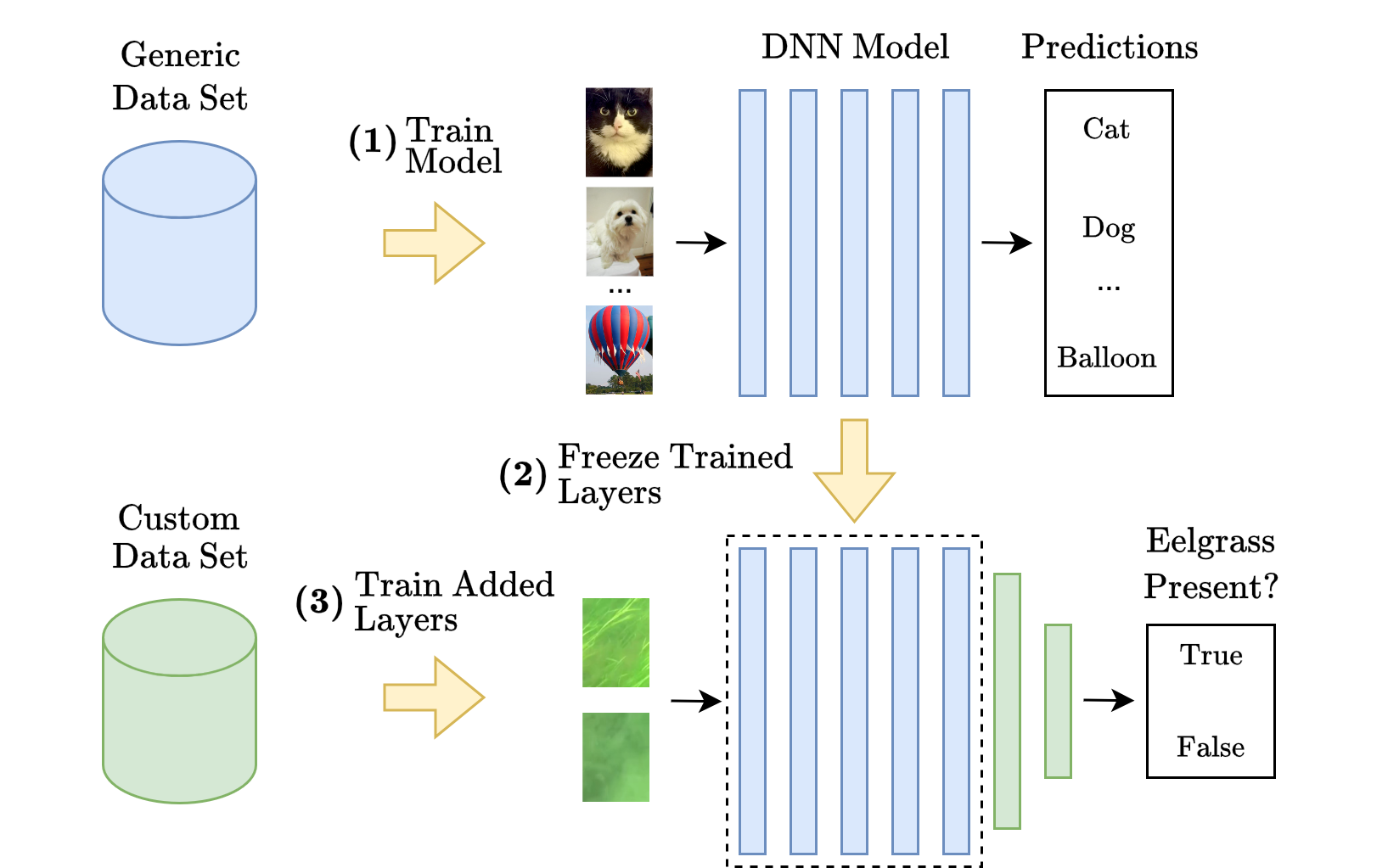}
    \caption{Transfer learning: (1) we use DNNs pre-trained on a generic data set (blue), here ImageNet \citep{russakovsky2015imagenet}. (2) The trained layers of each DNN model are frozen and infused into our custom training process. (3) We fine-tune a small number of additional layers on our eelgrass data set (green) to perform the target task of eelgrass detection.}
    \label{fig:transfer-learning}
\end{figure}

In this work, we adopt transfer learning to train a number of DNN models on the specific task of classifying images for two classes; ``Eelgrass Present'' and ``Eelgrass Absent''. In transfer learning, a DNN previously trained on a large generic data set, such as ImageNet \citep{russakovsky2015imagenet} for the task of classification, is reused for the same task but on a different, typically more specialized and smaller data set. For example, ImageNet-1k contains 1000 classes of animals and common objects, e.g. cats, dogs and cars. We are interested in the process of fine-tuning; instead of starting the DNN training process from scratch, we freeze the DNN layers trained to classify ImageNet, and add additional layers to the DNN architecture. By only training these additional layers on our specialized task of distinguishing between eelgrass absence and presence, we can reuse the DNN's knowledge and understanding of the generic task. Not only does this save computational resources, it also drastically improves learning efficiency and typically classification accuracy. Transfer learning is a popular approach in deep learning, to overcome data sparsity for specialized machine vision tasks. A visualization of the transfer learning mechanism can be seen in \cref{fig:transfer-learning}.
\looseness -1

One class of DNNs are convolutional neural networks. These contain convolutional layers that can learn features, like edges, corners and textures, by convolving multiple learnable filters across an input image. The output of these convolutional layers, a set of features maps, are the local responsiveness of an image area to each filter, while keeping the spatial relationships consistent. Convolutional neural networks are efficient at learning representative features, contributing to their effectiveness in image classification and object detection. However, they have inductive bias; they inherently assume that pixels in proximity have local dependencies that can be exploited during the learning process. This may not always be the case where a larger spatial integration is required for better contextual and spatial understanding. The current state-of-the-art DNN is the Vision Transformer model (ViT) \citep{dosovitskiy2020image}. These can overcome inductive biases by assigning a positional encoding to image patches and by incorporating a global attention mechanism, inspired by the latest advancements in natural language processing. 
\looseness -1

\review{We choose four DNN models in this work to analyze their ability to classify seagrass presence in images. We aim to test and analyze a range of state-of-the-art DNN architectures and model sizes. ResNet models are widely recognized for their strong performance capabilities in image classification tasks \citep{tan2018survey}. InceptionNetV3 is a well-established architecture and also referenced in prior work in seagrass analysis \citep{reus2018looking} and its popularity in image classification tasks. DenseNet architecture models show high accuracy in other image classification domains, such as medical imaging \citep{kim2022transfer}. Finally we choose Vision Transformers (configuration ViT-l-32) which are widely considered state-of-the-art in image classification. Based on our review, we are the first to apply ViTs in the context of eelgrass classification. Since our primary focus in this work is benchmarking model performance rather than proposing new architectures, we refer interested readers to the following works for in-depth introductions to convolutional neural networks and Vision Transformer architectures, as well as comparative evaluations in related domains such as medical imaging \citep{zhao2024review, taher2025large, takahashi2024comparison}. The selected architectures are summarized below;} 

\begin{itemize}[leftmargin=*]
    \item ResNet \citep{he2016deep}: a convolutional architecture incorporating residual blocks to address the challenge vanishing/exploding gradients in DNN training,
    \item InceptionNetV3 \citep{szegedy2016rethinking}, a convolutional architecture using multiple filter sizes to extract both local and global features, enhancing its capability to represent complex patterns in the input data,
    \item DenseNet \citep{huang2017densely}, a convolutional architecture which uses a dense connectivity pattern, allowing for better information and gradient flow throughout the network,
    \item Vision Transformer (ViT) \citep{dosovitskiy2020image}, a state-of-the-art vision model which incorporates an attention mechanism that overcomes inductive biases suffered from convolutional architectures.
\end{itemize}

All models are available online with pre-trained weights in PyTorch \citep{paszke2019pytorch}, and are pre-trained on the ImageNet \citep{russakovsky2015imagenet} dataset. We fuse each model with two additional fully-connected layers for performing classification. During the training process, we only fine-tune the weights of the additional layers, while the weights of each pre-trained model are frozen.

\begin{table}[!ht]
    \centering
    \caption{Model Hyperparameters: For each chosen model architecture we report the size of the two transfer learning (TF) layers for fine-tuning, the number of trained weights for fine-tuning, the number of frozen weights and the learning rate.}
        \begin{tabularx}{\columnwidth}{p{20mm}|X X X X}
        
        \thdr{Model} & \thdr{1st TF Layer} & \thdr{2nd TF Layer} & \thdr{\#Trained Weights} & \thdr{\#Frozen Weights} \\\hline
        ResNet & 512 & 512 & 131K & 11.4M  \\
        InceptionNetV3 & 512 & 256 & 1.2M & 25.1M  \\
        DenseNet & 512 & 256 & 623K & 18.1M  \\
        ViT & 512 & 256 & 513K & 306M  \\
        \end{tabularx}
    \label{table: inital model params}
\end{table}



\subsection{Eelgrass Coverage Estimation}
\label{sec:coverage-and-data-postprocess}

In a traditional eelgrass transect methodology eelgrass coverage is defined by having marine biologists annotate how much of the observed area is covered in eelgrass as a percentage. In Denmark eelgrass coverage is defined as ``the total substrate-specific coverage (determined) by projecting the outline of the foliage vertically onto the surface of the soft bottom and assessing the foliage's percentage coverage of the bottom.'' - translated from Danish \citep{bruhn2013au}. We find that this definition is unpractical to comply with when considering EIA video transects, where images are taken at an angle towards the seafloor. Alternatives such as visual coverage guides are highly subjective; the difference for example between 30\% and 55\% is visually hard to determine \citep{short2015seagrassnet}. 
\looseness -1

In this work we propose steps towards a novel, data-driven method for estimating eelgrass coverage. We are interested in developing a coverage estimation method that is based on the frequencies of outputs of the sample-based binary presence predictions of an image stream containing varying levels of eelgrass abundance. We bear in mind that our image data has certain characteristics; the images are taken at an angle towards the seafloor and thus the fields of view of two consecutive frames have spatial overlap. We speculate that we can therefore estimate the eelgrass coverage along the transect by calculating the temporal mean of the image stream with 0\% representing eelgrass absent and 100\% representing eelgrass present. The general expression of the temporal mean (TM) in a sequence of predictions $p$ in a range $r$ at instance $i$ is given by:

\begin{equation*}
\centering
\textrm{TM}_r (i) = \frac{p_i + p_{i+1} + p_{i+2} + \ldots + p_{i+r-1}}{r} = \frac{1}{r} \sum^{n=r-1}_{n=0} p_{i+n}.
\end{equation*}

We calculate the eelgrass coverage estimate with a range of $r=30$ samples, or 10 seconds of video footage. We deduce this rolling mean window size from the ship speed; assuming that the survey vessel sails at an average speed of around 1 m/s and each image contains around 1m\textsuperscript{2} area, then 10s of footage gives a spatial area of around 10m\textsuperscript{2}.

\begin{figure}[ht]
    \centering
        \begin{subfigure}[b]{\ifbool{singlecolumn}{0.45\textwidth}{\columnwidth}}
        \includegraphics[width=\linewidth, clip, trim=10mm 0mm 15mm 0mm]{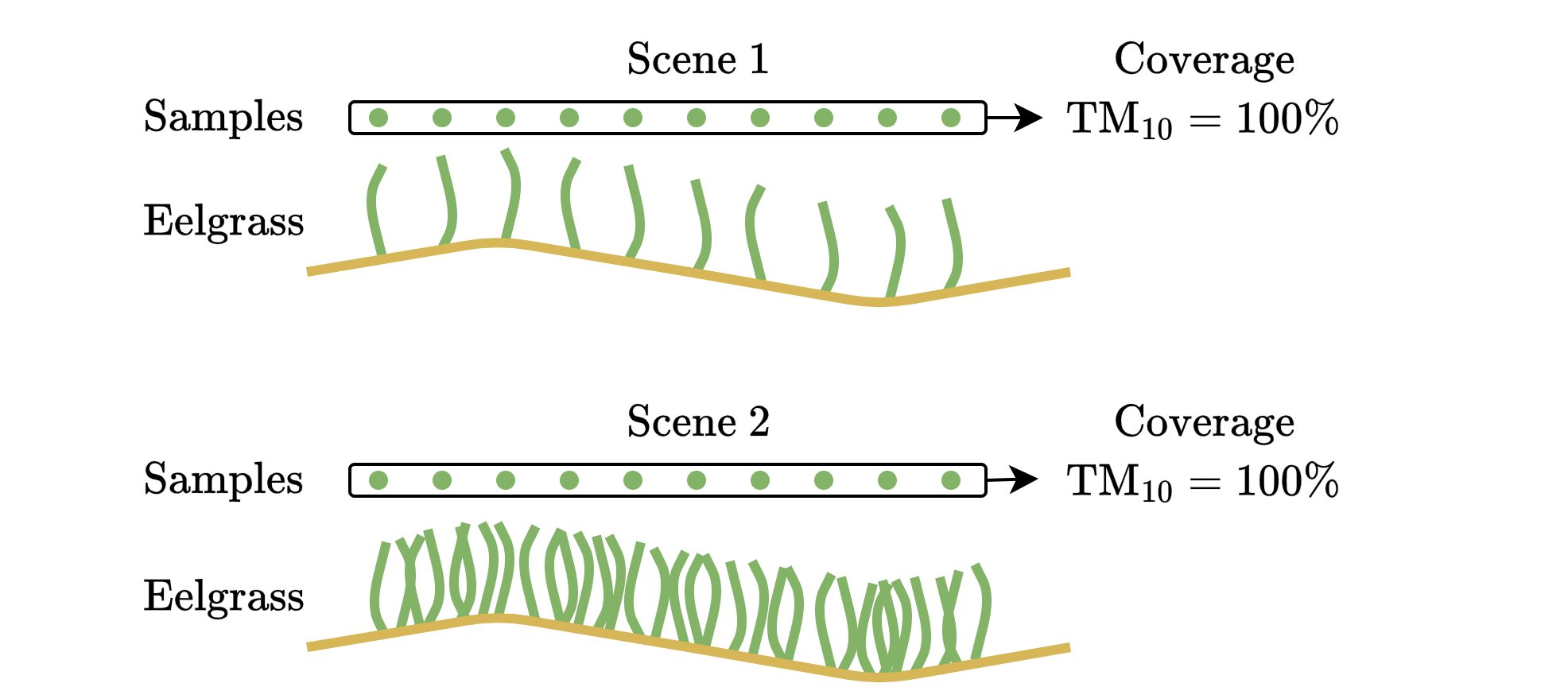}
        \caption{Example 1}
        \label{fig:limit-ex1}
    \end{subfigure}
    \begin{subfigure}[b]{\ifbool{singlecolumn}{0.45\textwidth}{\columnwidth}}
        \includegraphics[width=\linewidth, clip, trim=10mm 0mm 15mm 0mm]{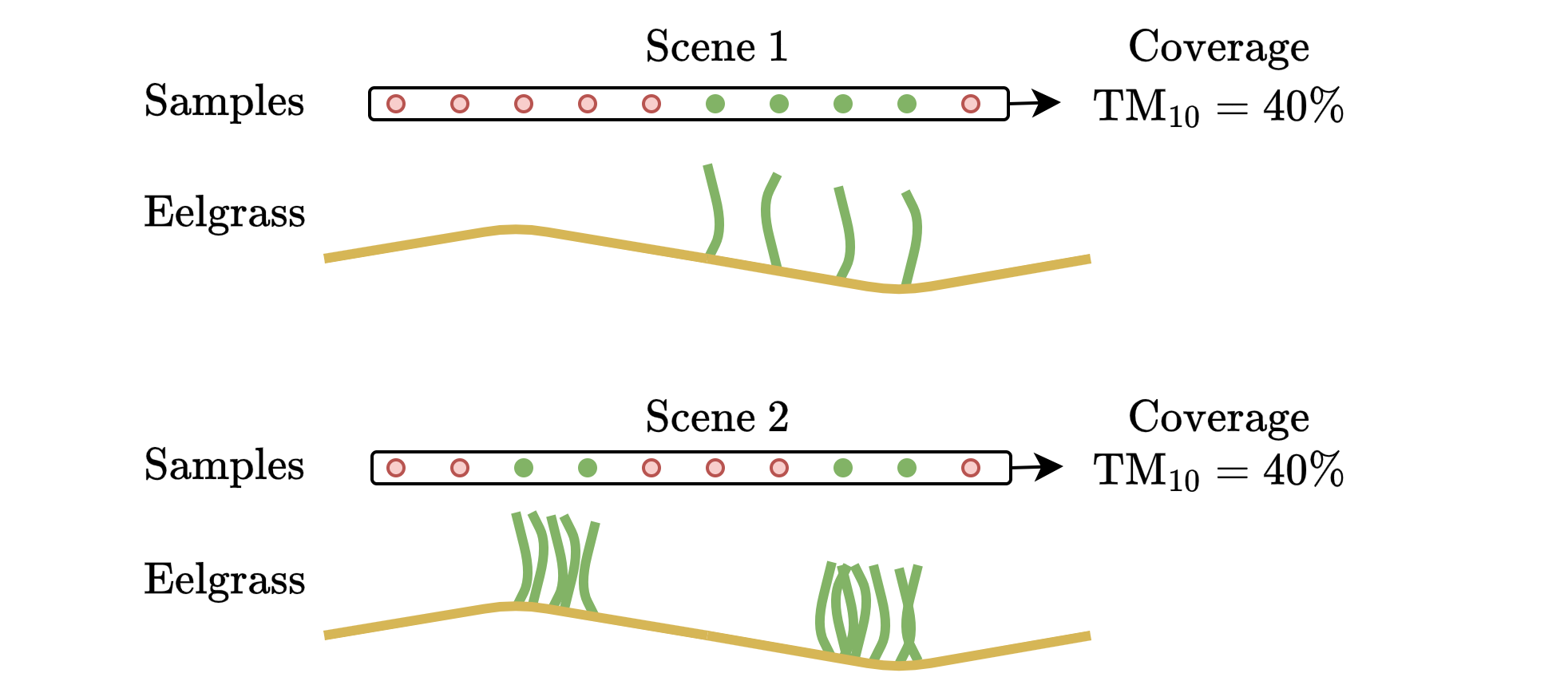}
        \caption{Example 2}
        \label{fig:limit-ex2}
    \end{subfigure}
    \caption{Limitations: even though the level of eelgrass abundance is different in the top scenes compared to the bottom scenes, eelgrass coverage estimation based on a temporal mean (TM) of ten samples gives the same result in both examples.}
    \label{fig:limits}
\end{figure}

We argue that this method can give an estimate of visual eelgrass coverage. For example, if there is a high frequency of consecutive frames with eelgrass present, then this denotes the presence of a consistent eelgrass coverage. If the frequency is low, then eelgrass becomes more patchy, and less frequent, denoting a lower eelgrass coverage. Our method has several advantages: (1) it is highly automated, scalable and practical once a suitable classifier is trained, (2) it can potentially predict eelgrass coverage in real-time and thus improve EIA transects monitoring even during their execution, (3) it is potentially less biased than human labeling and thus more reliable and repeatable.
\looseness -1

There are some limitations to our coverage estimation method as depicted in \cref{fig:limits}. First, consecutive images that have consistently little eelgrass present can give the same coverage estimation as consecutive images with abundant eelgrass (\cref{fig:limit-ex1}). Second, it may not be possible to determine the exact location of small and dense eelgrass patches as they give the same coverage estimation as larger sparse patches. The temporal mean obliterates this information (\cref{fig:limit-ex2}). Third, the method is sensitive to the temporal mean window size. Even though these limitations exist we argue that they are outweighed by the advantages of our method.
\looseness -1

To extract the overlaid positional meta-data from the transect videos we use the Tesseract Engine \citep{smith2007overview} for optical character recognition. We fuse the resulting GPS data with our eelgrass annotations and DNN-based predictions. We choose transects LYT-9, LYT-10, LYT-11, LYT-12, LYT-13, LYT-19 and LYT-20. For each transect, we use our ViT model to get a prediction for every 10th frame. Since all the videos are filmed at 30 frames per second, this gives three predictions per second. In total this creates 16796 predictions of eelgrass presence, related to a geographical position along each of the transects.
\looseness -1

\review{\section{Experimental Design}}
\label{sec:experiments}

\review{The overall objective for our experimental design is to identify the best performing and most robust DNN model out of our selection for eelgrass presence detection, such that we can subsequently employ its predictions for postprocessing and eelgrass coverage estimation. Since the quality of our training data is crucial we also conduct an annotation quality assessment. Our data strategy is shown in \cref{tab:dataset-overview}. We evaluate the performance of our chosen DNN models according to three metrics;}
\looseness -1

\vspace{0.2cm}

\begin{itemize}[leftmargin=*]
    \item accuracy ($\uparrow$): the percentage of correct predictions,
    \item area under receiver operating characteristic curve (AUROC or AUC) ($\uparrow$): ability of a model to distinguish between classes across all possible classification thresholds,
    \item test calibration error (CE) \citep{kumar2019verified} ($\downarrow$): the discrepancy between model predictions and the true probabilities given the model outputs. This is not a common metric for computer vision applications, however, it has been recently recommended for medical image analysis \citep{maier2024metrics}, a fine-grained classification task which shares properties with our data. CE scores range between 0 and 1, where $\textrm{CE}=0$ means the model is perfectly calibrated. 
\end{itemize}

\review{In this document, where higher values mean better performance, metrics are marked with ($\uparrow$) and where lower values mean better performance metrics are marked with ($\downarrow$).}

\review{\subsection{Annotation Quality Assessment}}
\review{We conduct an analysis of the annotations collected via SeagrassFinder AP since their correctness has a direct influence on the performance of our DNN model predictions. We calculate metrics to analyze the inter-annotator agreement level such as annotation agreement percentage, Cohen's Kappa for reliability of annotators as well as mistake rate. To estimate the mistake rate of annotators we consider two types of disagreements in annotations which we verify manually; a ``mistake annotation'' where a user accidentally clicks the wrong label button caused by an obvious lapse in judgment by visual inspection, and an ``ambiguous annotation'' where eelgrass is only partially visible or eelgrass presence was uncertain. By labeling ambiguous annotations we estimate the degree to which annotators make mistakes, and if the annotation task was not defined well enough. We additionally report the number of times an image is labeled and calculate the probability of agreement for each image normalized relative to the number of annotations.}


\review{\subsection{Initial Model Evaluation}
To gauge the baseline performance of each model architecture, we run an initial experiment where we choose LYT-20 as a test dataset, due to its frequent changing underwater landscape, caused by it's zig-zag constellation (see \cref{fig:transect-map}). For training, we use all other transects, except LYT-9. For these experiments we create the validation dataset by using a random split of a 20\% subset of images from the train dataset. The DNN classification layer configuration for the initial run (last layer and second last layer size) are configured as seen in \cref{table: inital model params}. For these runs we choose to configure the second last layer with more neurons than the last layer to guide the learning process towards a natural dimensionality reduction. All models are trained at a learning rate of $4.7\times 10^{-5}$, 33 epochs, batch size 22 and an early stopping mechanism with patience of 5 epochs.}


\review{\subsection{Leave-One-Transect-Out Validation} 
We choose a subset of models which performed best, and complete a cross-validation test. To overcome the challenge of imbalances in class distributions between different transects we devise an alternative approach to cross validation compared to traditional k-fold cross validation. We call this leave-one-transect-out cross validation. This approach only uses data from transects LYT-10, LYT-12 and LYT-20 due to their similar class distributions. On each cross validation fold we choose one of the three transects as the validation set, and the model trained on all other transects. We repeat this process of selecting one transect for validation three times, as such allowing for training and testing across all three transects. Upon completion we average the validation scores taken from all three experiments. This gives a good estimation of model performance, which does not directly overfit to one specific train and test dataset. The hyperparameters chosen for these experiments are the same as the hyperparameters used in initial experiments.}


\review{\subsection{Final Model Evaluation} 
Finally, based on the results of our ``leave-one-transect-out'' validation, we select the two highest performing models and complete the final model experiment. This involves using all transects except LYT-9 for training and validation and subsequently test on the LYT-9 transect.}
\looseness -1


\review{\subsection{Eelgrass Coverage Estimation} 
Since there are no expert labels available of eelgrass coverage estimation for the Lynetteholm EIA, we are limited to giving an intuition of the benefits of our novel estimation coverage. First, we visualize eelgrass presence predictions geographically alongside our eelgrass coverage estimations as a heatmap. To provide further evidence, we show a visual comparison between our approach and human annotations on a transect video from a different EIA in another location in Denmark (around 150km away). Here we have expert annotated coverage estimations. We cut a 15 minute section from this transect and show a visual comparison of our coverage predictions and expert labels in a time series.}


\section{Results}
\label{sec:results}

\review{In this section we present our results in the order of our data processing pipeline ``SeagrassFinder''. First we present our annotation quality analysis. Then we evaluate our chosen DNNs according to the task of binary classification of eelgrass presence and absence. Based on these results we choose the best-performing DNN and use its predictions for eelgrass coverage estimation.}

\review{\subsection{Annotation Quality Assessment}}

In total we have 19 different users creating more than 13000 annotations on more than 8900 different images. Generally, all annotators exhibit a high degree of annotation agreement, with annotators agreeing with one another on average 76\% of the time. Very few annotators (3 out of 19) exhibit a lower than 70\% agreement. We also investigate the impact of non-expert annotators annotating images and found that no bias is present between images annotated by non-expert annotators. There is also no connection between annotators with a lower agreement level, and their domain experience. This confirms that the classification of UW images for eelgrass presence is an annotation task humans are capable of, independent of annotators being experts or non-experts. This is still true for more challenging images with high levels of blur, color-distortion or low contrast. 
\looseness -1 

\begin{figure}[!ht]
    \centering
    \begin{subfigure}[b]{\ifbool{singlecolumn}{0.35\textwidth}{\columnwidth}}
        \includegraphics[width=\linewidth]{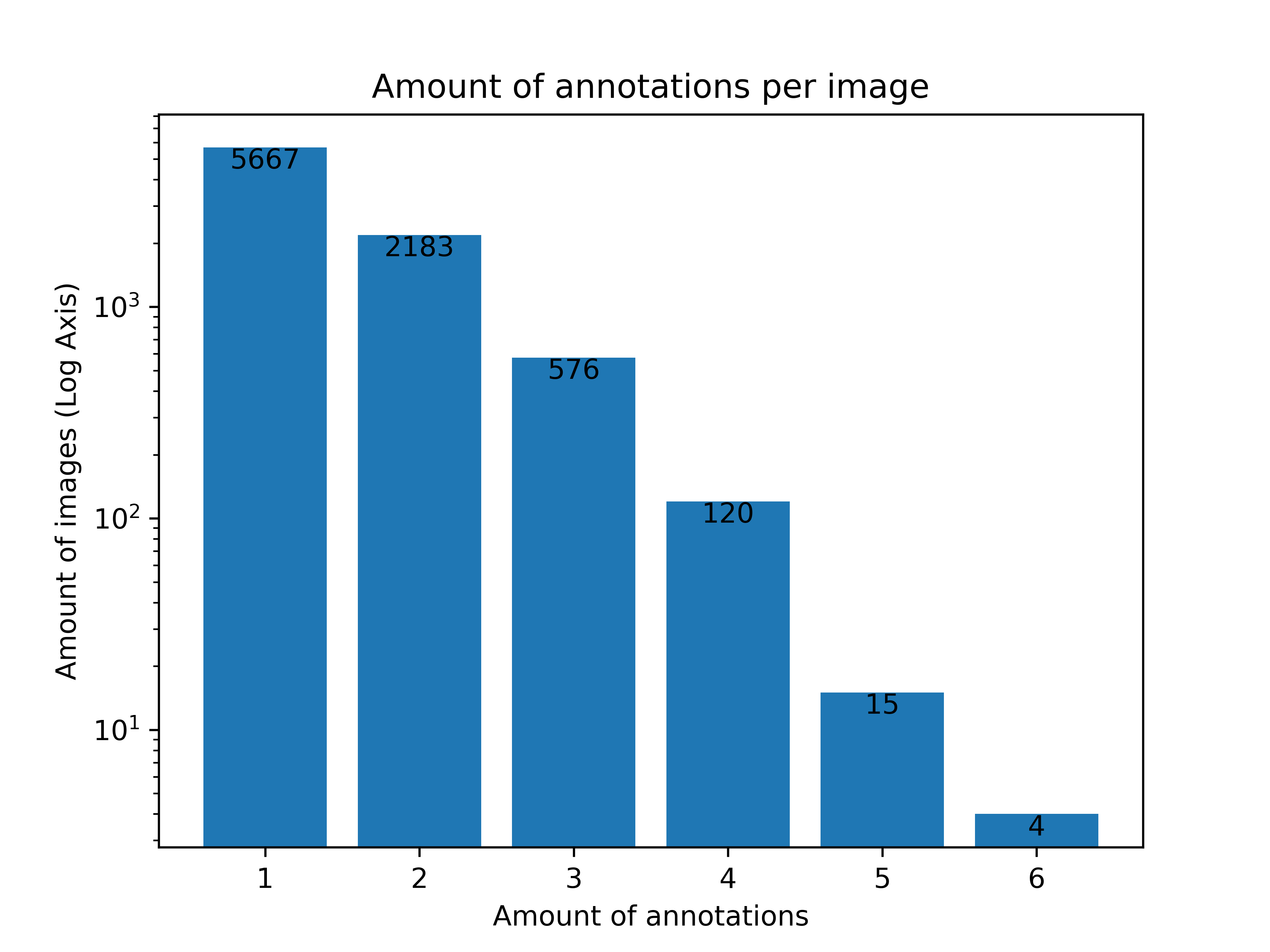}
        \caption{}
        \label{fig:number-agreements}
    \end{subfigure}
    \begin{subfigure}[b]{\ifbool{singlecolumn}{0.35\textwidth}{\columnwidth}}
        \includegraphics[width=\linewidth]{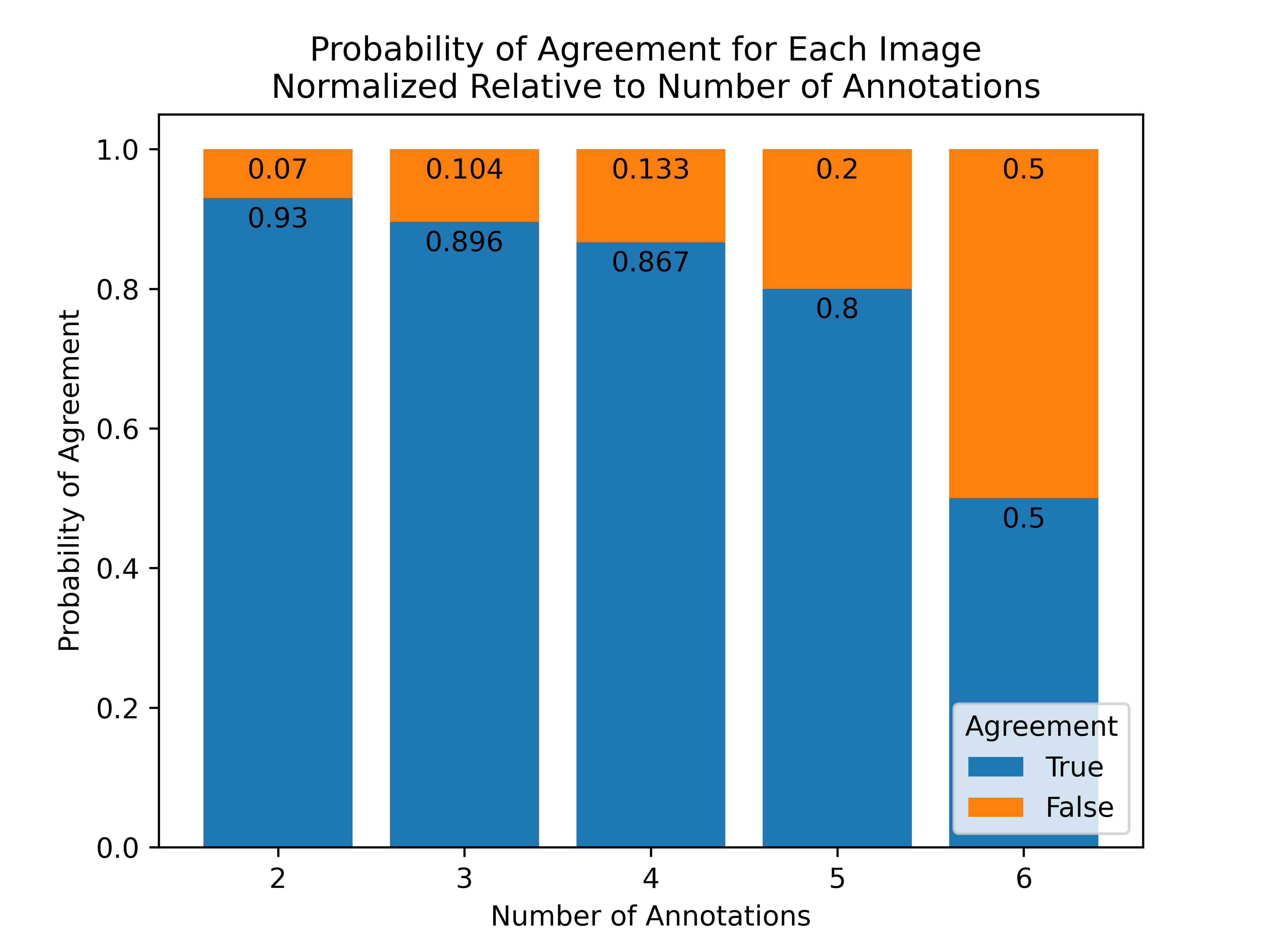}
        \caption{}
        \label{fig:prob-agreement}
    \end{subfigure}
    \caption{(a) The number of images on log-scale vs. the number of times the image was annotated. (b) The probability of agreement for each number of annotations.}
\end{figure}

\Cref{fig:number-agreements} shows the number of times an image is labeled. The probability of agreement for each image normalized relative to the number of annotations can be seen in \cref{fig:prob-agreement}. For images that are annotated twice, the annotators achieve an overall 93\% agreement. Cohen's Kappa scores are calculated between every annotator's subset of common annotated images and reported in the format of a matrix (see appendix \cref{fig:cohens-kappa-scores}). Using magnitude guidelines defined by Landis et al. \citep{landis1977measurement} we find that on average there is a level of agreement that is of \textit{substantial} to \textit{almost perfect} magnitude. On occasions where less than or moderate agreement is seen, we find that this is caused by too small a sample size of commonly annotated images. 
\looseness -1

Based on the ratio of ``ambiguous'' to ``mistake'' annotations we dismiss the presumption that the annotation task was too vague. However it does highlight that the task of annotating can be at times very difficult to complete. Among the annotators, mistake annotations occur at a fairly low rate, on average at around 8\% (see \cref{fig:mistake-per-user}.) There seems to be no relation between a user's percentage level of disagreement and their mistake percentage.
\looseness -1


\review{\subsection{Initial Model Evaluation}}

The results of our initial model evaluation experiments are shown in \cref{table:init-results}. Based on the high level of AUROC scores, and a low variation for all models, we ascertain that the task of eelgrass classification can in fact be conducted by DNN models. Secondly, the models trained and tested on UW-enhanced images show a higher accuracy compared to their non-enhanced counterparts in 26/30 cases. Thus UW-enhancement has a positive effect on the models' ability to predict eelgrass presence accurately in the context of transfer learning with pre-training on Imagenet. We conclude that some models may benefit from UW-image enhancement. Generally we find that the Vision Transformer and DenseNet architecture show good performance on the target task. Surprisingly the smallest ResNet model, ResNet-18\texttt{++}, gives the best results across all three metrics among the ResNet models. The Vision Transformer shows the best results overall compared to all DNN architectures across all three metrics.
\looseness -1

We are interested in how the test calibration error changes between the top AUROC performing models. This highlights that different models are over/under-confident during prediction. For example DenseNet-169\texttt{++} and InceptionNetV3\texttt{++}, which both have equal AUROC scores, however DenseNet-169\texttt{++} has a higher test calibration error than InceptionNetV3\texttt{++}. We conclude that InceptionNetV3 is a more stable model, and not as prone as to over/under-predicting the posterior probabilities compared to the DenseNet model.

\begin{table}[h]
    \centering
    \caption{Test results from initial experiments, with models trained on LYT-5, LYT-10, LYT-12 and LYT-14 and tested on LYT-20. The 3 metrics of interest are test accuracy, are under ROC curve (AUROC) and test calibration error (CE). Higher scores mean better performance and are marked by ($\uparrow$) and vice versa marked by ($\downarrow$). Models trained and tested on UW-enhanced images are marked by \texttt{++}.}
    \begin{tabular}{l|ccc}
        \thdr{Model} & \thdr{Accuracy ($\uparrow$)} & \thdr{AUROC ($\uparrow$)} & \thdr{CE($\downarrow$)} \\\hline \\[-3mm]
        ResNet-18               & 0.769 & 0.892 & 0.203 \\
        ResNet-18\texttt{++}    & \textbf{0.792} & \textbf{0.902} & \textbf{0.188} \\
        ResNet-34               & 0.730 & 0.877 & 0.252 \\
        ResNet-34\texttt{++}    & 0.777 & 0.876 & 0.215 \\
        ResNet-50               & 0.784 & 0.895 & 0.191 \\
        ResNet-50\texttt{++}    & 0.766 & 0.894 & 0.222 \\
        ResNet-101              & 0.767 & 0.887 & 0.211 \\
        ResNet-101\texttt{++}   & 0.782 & 0.892 & 0.197 \\
        ResNet-152              & 0.748 & 0.886 & 0.229 \\
        ResNet-152\texttt{++}   & 0.761 & 0.895 & 0.220 \\ \hline \\[-3mm]
        InceptionNetV3          & 0.794 & 0.893 & 0.184 \\
        InceptionNetV3\texttt{++} & \textbf{0.828} & \textbf{0.909} & \textbf{0.155} \\ \hline \\[-3mm]
        DenseNet-121            & 0.779 & 0.899 & 0.213 \\
        DenseNet-121\texttt{++} & 0.813 & \textbf{0.914} & 0.176 \\
        DenseNet-169            & 0.788 & 0.902 & 0.208 \\
        DenseNet-169\texttt{++} & 0.826 & 0.908 & 0.160 \\
        DenseNet-201            & 0.818 & 0.907 & 0.169 \\
        DenseNet-201\texttt{++} & \textbf{0.835} & \textbf{0.914} & \textbf{0.156} \\ \hline \\[-3mm]
        ViT                     & 0.830 & 0.914 & 0.161 \\
        ViT\texttt{++}          & \textbf{0.867} & \textbf{0.938} & \textbf{0.125} \\

    \end{tabular}
    \label{table:init-results}

\end{table}


\review{\subsection{Leave-One-Transect-Out Validation}}

As can be seen in \cref{table: cross validation results}, the ViT architecture continuously shows the best performance, with the enhanced version ViT\texttt{++} performing best. Using models on UW-enhanced data also continues to provide the highest model scores across all model architectures. We also notice that the smaller DenseNet-169\texttt{++} model has a higher accuracy than the DenseNet-201, although performing worse in the AUROC and CE categories. This highlights the importance of using multiple different metrics to evaluate models. 
\looseness -1

\begin{table}[!ht]
    \centering
    \caption{Leave-one-transect-out cross validation results. The 3 metrics of interest are test accuracy, area under ROC curve (AUROC) and test calibration error (CE). Higher scores mean better performance and are marked by ($\uparrow$) and vice versa marked by ($\downarrow$). Models trained and tested on UW-enhanced images are marked by \texttt{++}.}
    \label{table: cross validation results}
    \begin{tabular}{l|ccc}
        \thdr{Model} & \thdr{Accuracy ($\uparrow$)} & \thdr{AUROC ($\uparrow$)} & \thdr{CE ($\downarrow$)} \\\hline
        ResNet-18                   & \textbf{0.870} & \textbf{0.939} & 0.098 \\
        ResNet-18\texttt{++}        & 0.867 & 0.936 & \textbf{0.097} \\\hline
        InceptionNetV3              & 0.855 & 0.936 & 0.115 \\
        InceptionNetV3\texttt{++}   & \textbf{0.868} & \textbf{0.940} & \textbf{0.108} \\\hline
        DenseNet-169                & 0.866 & 0.942 & 0.105 \\
        DenseNet-169\texttt{++}     & \textbf{0.879} & 0.943 & 0.099 \\
        DenseNet-201                & 0.874 & 0.944 & 0.105 \\
        DenseNet-201\texttt{++}     & 0.875 & \textbf{0.946} & \textbf{0.095} \\\hline
        ViT                         & 0.852 & 0.938 & 0.126 \\
        ViT\texttt{++}              & \textbf{0.901} & \textbf{0.953} & \textbf{0.082} \\
    
    \end{tabular}

\end{table}


\review{\subsection{Final Model Evaluation}}

We report our final model test results in \cref{table:final-results}. In the final model training and testing, for non-enhanced models both DenseNet-201 and Vision Transformer achieve high scores, with both models gaining high AUROC scores on the test set. Both models also achieved very low calibration error scores, 0.084 and 0.086 respectively, showing both models capability to not only predict accurately, but also confidently. Overall the Vision Transformer performed best across all metrics. Both models showed an increase in performance when trained and tested on the UW-enhanced images. With a total training time of around 30 minutes we show that our transfer learning approach is highly efficient, compared to training a ViT from scratch which can take up to three days on four GPUs \citep{touvron2021training}.
\looseness -1

\begin{table}[ht]
    \centering
    \caption{Final Model Results: Models trained and tested on enhanced images are marked by \texttt{++}.}
    \begin{tabularx}{\columnwidth}{l|XXXp{1.4cm}}
        \textbf{Model}  & \textbf{Accuracy ($\uparrow$)} & \textbf{AUROC ($\uparrow$)} & \textbf{CE ($\downarrow$)} & \textbf{Train Time ($\downarrow$)} \\\hline
        DenseNet-201            & 0.877 & \textbf{0.955} & 0.113 & 30min \\
        DenseNet-201\texttt{++} &\textbf{0.878} & \textbf{0.955} & \textbf{0.110} & 30min \\ \hline
        ViT                     & 0.877 & 0.955 & 0.113 & 28min \\
        ViT\texttt{++}          & \textbf{0.902} & \textbf{0.959} & \textbf{0.087} & 28min \\

    \end{tabularx}
    \label{table:final-results}
\end{table}


\review{\subsection{Eelgrass Coverage Estimation}}

\begin{figure}[ht]
    \centering
        \begin{subfigure}[b]{\ifbool{singlecolumn}{0.45\textwidth}{0.48\columnwidth}}
        \centering
        \includegraphics[width=\linewidth]{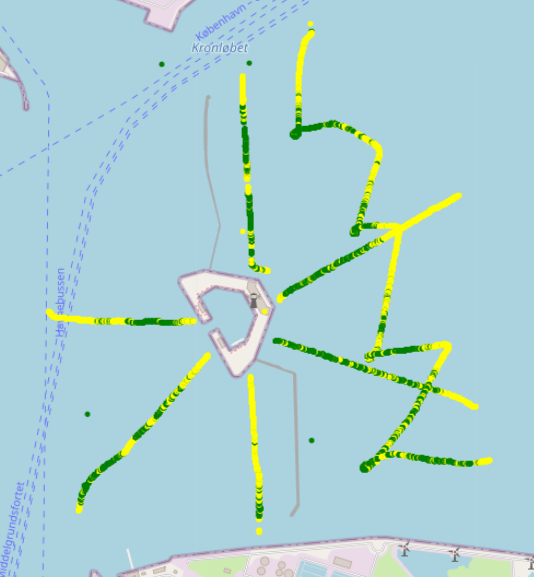}
        \caption{}
        \label{fig:eelgrass-presence-mapped}
    \end{subfigure}%
    \ifbool{singlecolumn}{}{\hfill}
    \begin{subfigure}[b]{\ifbool{singlecolumn}{0.45\textwidth}{0.48\columnwidth}}
        \centering
        \includegraphics[width=\linewidth]{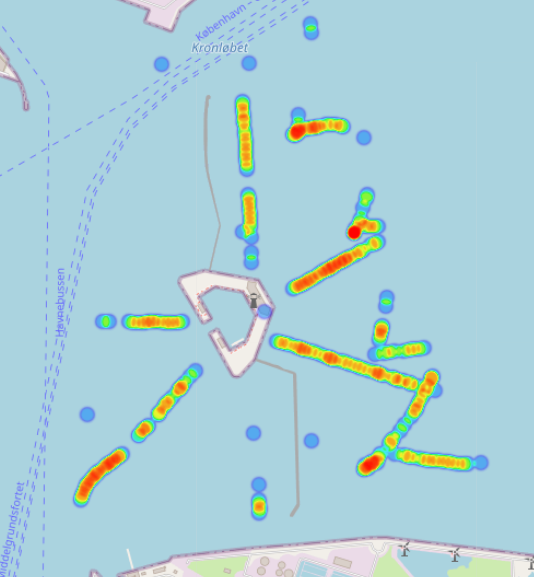}
        \caption{}
        \label{fig:eelgrass-presence-heatmap}
    \end{subfigure}%
    \caption{(a) Eelgrass presence: the lines mark transect surveys conducted along planned survey paths, green dots indicate detected eelgrass presence along each transect, yellow dots indicate eelgrass absence. (b) Eelgrass coverage: an estimation of eelgrass coverage in the benthic environment represented by a heat map of the transect surveys.}
  
\end{figure}

We map the ViT predictions from LYT-9, LYT-10, LYT-11, LYT-13, LYT-19, LYT-20 geographically in \cref{fig:eelgrass-presence-mapped}. The green marks denote the presence of eelgrass, and yellow the absence. In \cref{fig:eelgrass-presence-heatmap} all eelgrass present data points are mapped onto a heat map, giving us a good visual estimation for eelgrass coverage in the area.
\looseness -1

\begin{figure}[ht]
    \centering
    \includegraphics[width=\ifbool{singlecolumn}{0.6\linewidth}{\linewidth}]{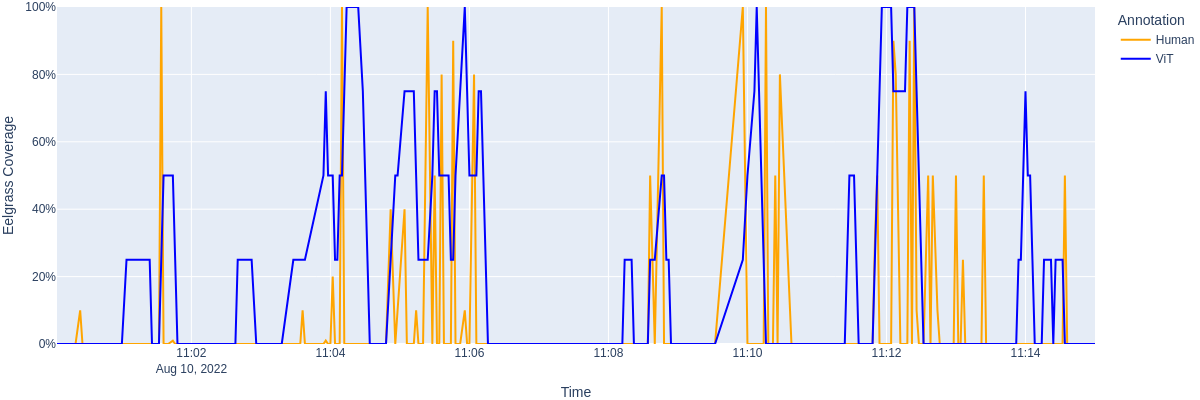}
    \caption{Visual Coverage Estimation: (blue) our new approach based on the temporal mean of ViT predictions of a transect video, (orange) human annotated eelgrass coverage.}
    \label{fig:vit-preds-vs-annotations}
\end{figure}

Since we do not have expert labels on eelgrass coverage available of the LYT transects, we show how our coverage estimation compares with manual labels of a different EIA transect video. In \cref{fig:vit-preds-vs-annotations} we show the eelgrass coverage estimation based on our ViT predictions compared to video analysis labeled by experts. The respective transect video was taken at a location around 150km away from the LYT video transects. \review{\Cref{fig:vit-preds-vs-annotations} serves as a qualitative demonstration of how our coverage estimation method captures ecologically meaningful patterns in an automated and efficient manner. \Cref{fig:vit-preds-vs-annotations} also shows that our method is able to generalizable to an environment with a substantially different seabed. Additionally, we computed a Spearman correlation test for the two time series which is $0.3169$ with a p-value of $2.2331 \times 10^{-8}$. However it is important to note, that expert labels are sparse, highly subjective and often not repeatable, therefore this result should be interpreted with caution as there is no baseline for what constitutes a ``good'' value.  Selecting an appropriate metric and establishing its utility for experts is an avenue for further research.}

\section{Discussion}
\label{sec:discussion}
Here we discuss the ecological importance of our findings, as well as limitations, future work and applications.
\looseness -1


\subsection{Video Data Limitations and Solutions}

During the sailing of the transect surveys, the recorded video footage is streamed to video processing software which adds GPS location, water depth and heading (i.e. course over ground measured in degrees) to each video as an overlay. Streaming lag can cause a delay in the timestamps in the imprinted overlay. This is problematic when trying to align the images to GPS locations during post-processing. Additionally there is a mismatch ($\sim$ 5-20m) between the imprinted GPS locations of the vessel and the actual camera position on the towed sled. One approach to improve GPS position accuracy, and generally make steps towards automated eelgrass inspection, is to deploy more advanced multi-purpose Remotely Operated Towed Vehicles (ROTVs) or Autonomous Underwater Vehicles (AUVs). Compared to our basic mechanical towed sled, ROTVs offer many advantages; they can carry higher payloads and thus multiple onboard sensors. If fused together, the data streams of these sensors can provide more detailed and accurate insights of the benthic environment of transect surveys.

\review{\subsection{Automation of Seagrass Monitoring}
The automation of seagrass detection and coverage estimation is crucial for enabling efficient, timely and consistent monitoring of the health of underwater ecosystems. By eliminating the need for manual annotation, automated methods allow for processing of high volumes of video data and thus enable us to detect and react to ecological changes in ecosystems and respond more rapidly for example through restoration efforts by transplantation \citep{waycott2009accelerating}. Repeated and data-driven monitoring also enables the assessment of environmental impacts associated with coastal developments such as tunnels or windfarms, that is, before, during and after construction. Since seagrass meadows also act as nursery grounds for fish, it is estimated that 20\% of the global fishery productivity is provided by seagrass meadows \citep{unsworth2019seagrass}. As such, automated and repeated seagrass monitoring can potentially serve as an indirect indicator of local fish stock health. Automated seagrass coverage estimation could also serve as a valuable addition to blue carbon strategies and aid in tracking climate mitigation targets \citep{unsworth2019global, fourqurean2012seagrass}. While seagrass detection in a benthic environments provides information about the presence and locality of seagrass meadows, tracking seagrass coverage offers more nuanced insights into their condition, density and health. Activities like dredging, sedimentation, or eutrophication may reduce eelgrass density before total loss and is therefore an important advancement for marine conservation and coastal management. Here we would like to emphasize the importance of drawing a clearer distinction between spatial seagrass coverage, which refers to its actual density of vegetation, and pixel-based coverage, which simply measures the proportion of image pixels identified as seagrass. To the best of our knowledge, this distinction is currently overlooked in computer vision approaches to seagrass coverage estimation.} 
\looseness -1

\subsection{Automated Transect Surveys with AUVs}
Deploying Autonomous Underwater Vehicles (AUVs) has multiple benefits for automated transect surveying compared to a towed sled. An AUV with a forward-facing camera can perform EIAs autonomously, without the need for a manual operator. The AUV might have the high-level objective of inspecting seagrass meadows and therefore needs to detect the meadow boundaries in real-time, as for example demonstrated by Ruscio et al. \citep{ruscio2023autonomous}. Our analysis of DNN performance enables robotics engineers to now choose a suitable DNN according to their specific hardware constraints, and integrate a trained DNN into the AUV navigation stack for an improved inspection mission protocol. Simply increasing the operating velocity of the AUV in areas devoid of seagrass can cause substantial cost reductions, particularly in the rental of research vessels. In addition to this, AUVs are able to perform more advanced missions than simple transect lines. A typical inspection mission can be conducted in form of a lawnmower pattern, which covers a much larger area and thus gives a better understanding of the health of the underwater environment.
\looseness -1


\subsection{Eelgrass Coverage Estimation}

In this work, we present a novel idea for eelgrass coverage estimation to overcome practical constraints of traditional methods. We outline the benefits and limitations of our approach in \cref{sec:coverage-and-data-postprocess}. Additionally we argue, that human labeling inherently suffers from observer bias. Although DNNs are not immune to biases, they can reduce variability across images and perspectives as well as offer a consistent and deterministic analytical process, once they are trained. Human labeling also suffers from the tendency of annotators to focus disproportionately on specific parts of an image. This issue is particularly pronounced here since our images are captured at an angle towards the seafloor; eelgrass in the foreground might induce the annotator to indicate a high coverage percentage even if there is little eelgrass present in the background. We assert that our data-driven approach of calculating the temporal mean of our DNN-based binary predictions offers a robust and accurate approach to estimating eelgrass coverage in the wild. Here we would like to stress that this approach should act as a supportive tool for marine biologists to conduct EIAs in a more data-driven manner. The tools presented in this work should not be seen as a replacement for expert domain knowledge, rather, a showcase to apply state-of-the-art technologies to support marine experts in their work. With our methodology, marine experts can finally focus their efforts on interpretation of environmental impact instead of data annotation. This increases quality assurance, but also employee well-being by reducing time spent on tedious manual tasks. Additionally, we suspect that our methodology and analytic setup can easily be expanded to automatically detect other taxonomically related species of submerged aquatic vegetation with similar visual characteristics.
\looseness -1


\subsection{Data Availability and Underwater Image Quality}
There is a huge growth in interest for underwater image analysis, not only for EIAs, but also in AUV camera footage and static underwater cameras. However, this discipline has many hurdles. First, the limited availability of relevant and diverse data sets poses a hard problem, not only for marine biologists but also researchers in computer vision. Second, the amount of financial resources required to commission a research vessels can be extremely high. Even for static camera setups underwater, we require advanced technologies for efficiently managing battery power and protecting hardware from water damage. To overcome the sparsity of labeled data sets, Abid et al. \citep{abid2024seagrass} propose unsupervised curriculum learning to address uncertainties in seagrass classification. Here, the learning algorithm chooses image samples in order of difficulty, from easiest to hardest, for a more robust learning strategy on the DeepSeagrass data set \citep{raine2020multi}.

The visual quality of underwater images taken in the wild poses another big challenge. Some visual aspects such as back scatter, haze effects, caustics, low light and low contrast can lead to uncertainties in the DNN predictions and potential misclassifications. In the context of transfer learning in medical images, recent work has shown that the choice of the source data for DNN pretraining has an effect on the performance of the subsequent target task \citep{juodelyte2024source}. We are curious whether selecting an alternative domain dataset to ImageNet for pretraining could mitigate the domain shift challenges observed from on-land to underwater images. This investigation may improve DNN accuracy and generalization, however this remains out of scope.
\looseness -1

During visual inspection of the transect videos, we noticed motion blur in some images. This was initially attributed to the camera's low quality and frame-rate and later suspected to stem from an automated video compression. To overcome this we recommend investing in a high-quality waterproof camera system. It is important to mention that humans are good at filtering out such motion trails, while DNNs are not, due to their dependence on raw pixel data. Therefore this should be taken into consideration when using this dataset for future applications. 

\subsection{Underwater Image Enhancement}
We find that applying Deep WaveNet \citep{sharma2023wavelength} as an UW image enhancement tool proves to be effective in improving model accuracy. However, the computational demands are substantial, requiring approximately three hours of processing time on an A10G GPU for the entire training set, or around 1.2 seconds per image. The trade-off between the computational investment and the incremental performance gains must be carefully considered in practice, given the rising carbon footprint of training of DNNs \citep{luccioni2023counting}. Additionally, it is also important to consider the potential for the Deep WaveNet model to introduce bias in eelgrass presence classification based on water depth. As Sharma et al. show, the receptive field size of global features is more effective at greater water depths, than in shallow waters. Since all transects used for model training and testing occur in water depths of 2-14m, the impact of Deep WaveNet in this context is minimal but cannot be entirely discounted. 


\section{Conclusion}
\label{sec:conclusion}

This work shows the potential of deep neural networks to automate the process of seagrass detection in underwater videos from challenging visual conditions in the wild. By training deep neural networks to classify underwater images into ``Eelgrass Present'' and ``Eelgrass Absent'' classes, our proposed methodology allows for the efficient processing of large volumes of video data collected in environmental impact assessments, enabling the acquisition of much more detailed information on seagrass distributions compared to current manual annotation methods. Our results show that deep neural networks, particularly Vision Transformers, can achieve high performance in predicting eelgrass presence, with AUROC scores exceeding 0.95 on the final test dataset. We also show an effective implementation of transfer learning, and that the application of the underwater image enhancement model further improves models' capabilities.
\looseness -1

Beyond the technical aspects, we also propose a methodology which combines existing data acquisition methods like video transects with a streamlined annotation platform, as well as advancements in the field of machine learning. We highlight how such a methodology has significant benefits, including faster and more accurate data annotation, the potential for automated quality assurance and long term automation benefits.  We emphasize that deep neural networks are positioned as supportive tools to augment the work of marine biologists. Broadly we prospect that this work serves as a framework for a marine imaging workflow which can be implemented on other underwater flora. 
\looseness -1

Overall, this project demonstrates the value that deep learning can bring to the field of marine ecology and environmental monitoring. By automating the processing of underwater video data, it opens up new possibilities for gaining a deeper understanding of seagrass ecosystems and their response to environmental changes. As the human impacts on seagrass beds continue to increase worldwide, tools like the one we propose, will become increasingly important for effective conservation and management efforts.  

\section{Acknowledgments}

The authors would like to thank their collaborators at DHI: Aron Lank Jensen, Jesper Goodley Dannisøe and Lindsey Aies. We would like to thank ``By \& Havn'' for kindly allowing us to use and share their data, a crucial resource for our findings and the research community. This project was funded and run in collaboration with DHI \citep{dhi_dhi_2023}. This project has received funding from the European Union’s Horizon 2020 research and innovation programme under the Marie Skłodowska-Curie grant agreement No 956200.

\section{Data Availability}
The data used to support our findings is titled ``SeagrassFinder: An Underwater Eelgrass Image Classification Dataset'' and is publicly available on zenodo:\\ \href{https://zenodo.org/records/13904604}{https://zenodo.org/records/13904604}.


\printcredits


\appendix

\section{Appendix}

\subsection{Annotation Analysis}
\begin{figure*}[!htbp]
    \centering
    \includegraphics[width=\ifbool{singlecolumn}{0.8\textwidth}{0.6\textwidth}]{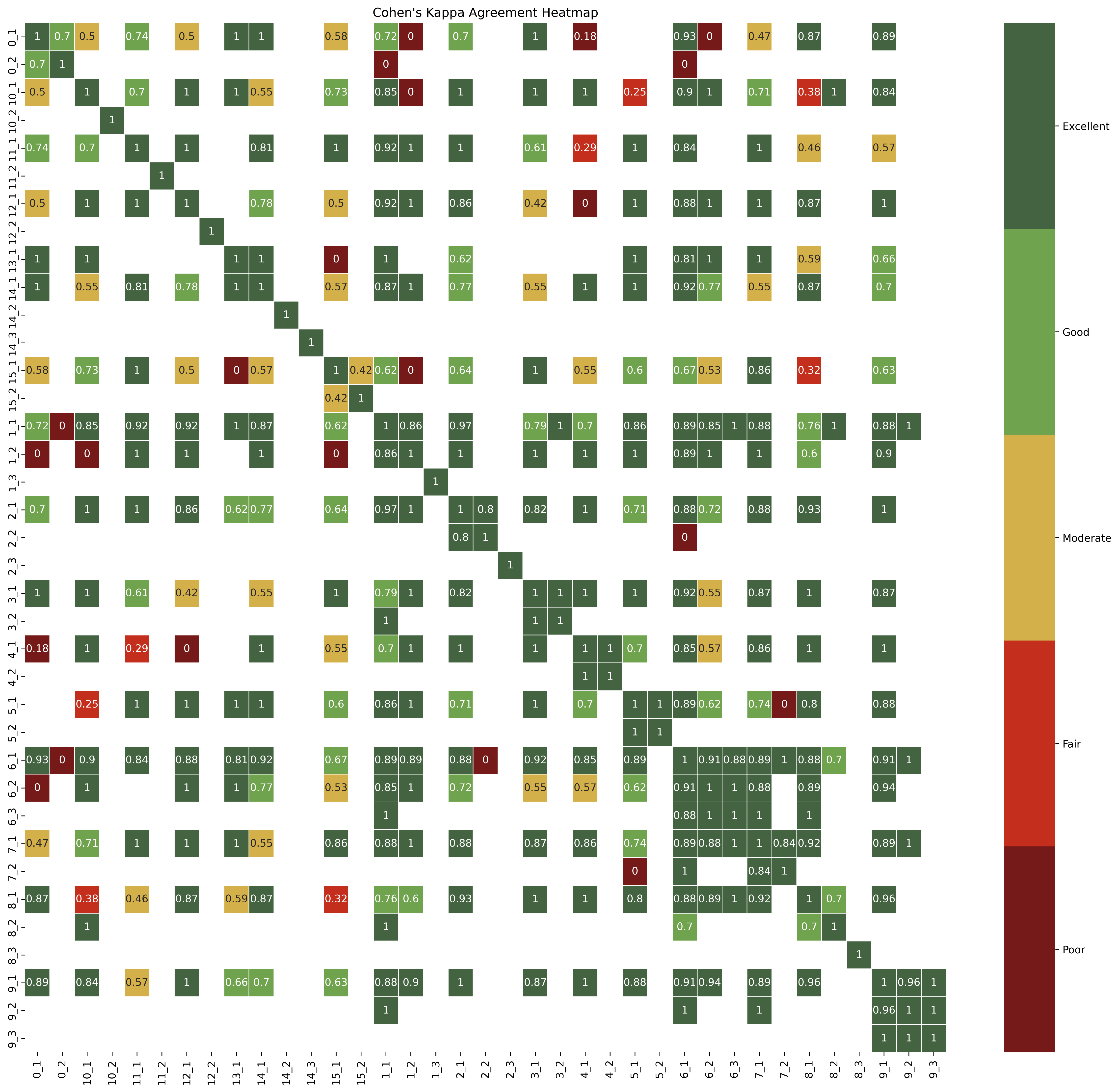}
    \caption{Cohens kappa scores for all annotators}
    \label{fig:cohens-kappa-scores}
\end{figure*}

\begin{figure}[!ht]
    \centering
    \includegraphics[width=\ifbool{singlecolumn}{0.6\textwidth}{\columnwidth}]{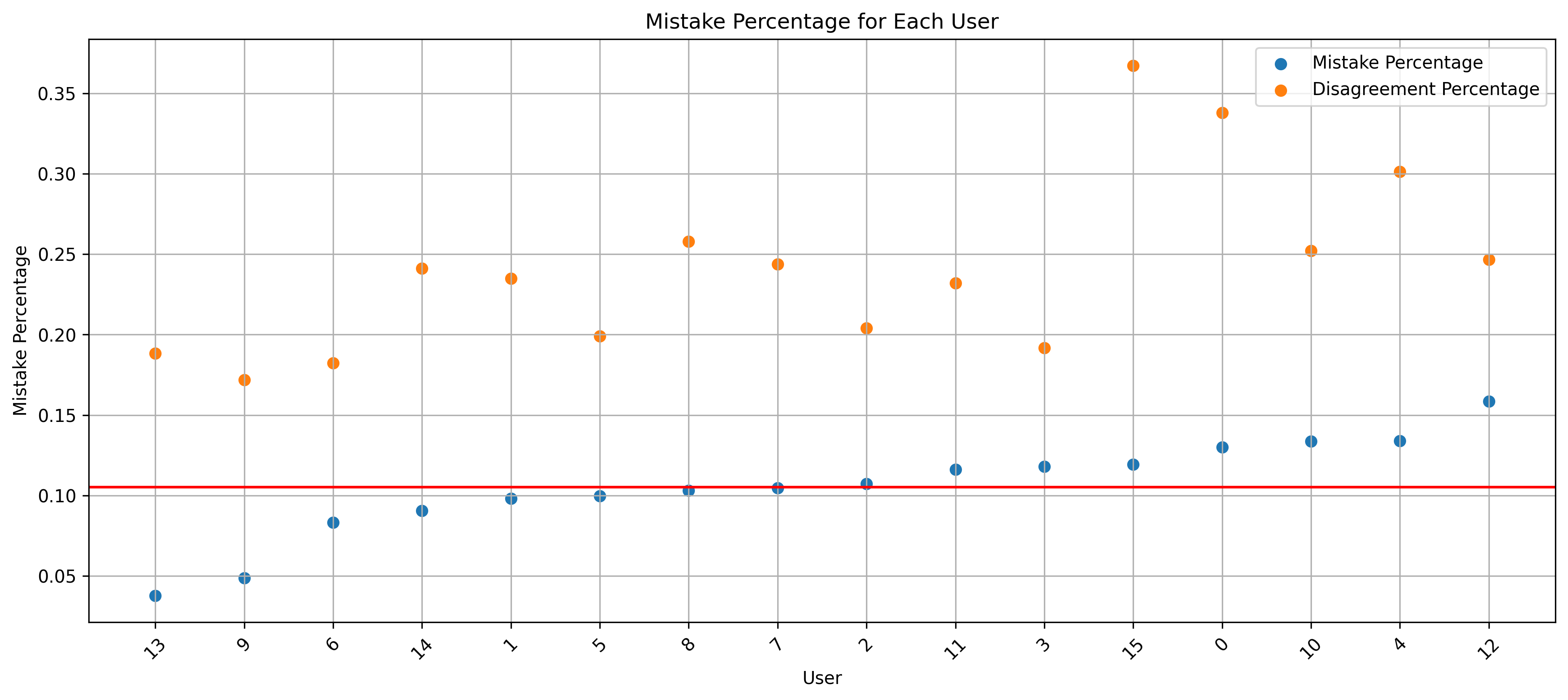}
    \caption{Mistakes percentage and disagreement percentage for each annotator.}
    \label{fig:mistake-per-user}
\end{figure}

We report Cohens kappa scores for all annotators in \cref{fig:cohens-kappa-scores} and mistake percentage and disagreement percentage in \cref{fig:mistake-per-user}.
\looseness -1

\subsection{Glossary of Acronyms}
We provide a glossary with all acronyms used in our work for reference in \cref{tab:glossary-acronyms}.

\begin{table}[h]
    \centering
    \caption{A list of all acronyms used in this work in alphabetical order.}
    \setlength{\tabcolsep}{5pt} 
    \begin{adjustbox}{center}
    \begin{tabular}{l|l}
        \hline
        \textbf{Acronym} & 
        \parbox{1cm}{Description} \\ \hline
        AP & Annotation Platform \\
        AUV & Autonomous Underwater Vehicle \\
        AUROC & Area Under Receiver Operating Characteristic Curve \\
        CE & Calibration Error \\
        CMOS & Complementary Metal Oxide Semiconductor\\
        DNN & Deep Neural Network \\
        EIA & Environmental Impact Asessment \\
        FPS & Frames Per Second \\
        GPS & Global Positioning System \\
        GPU & Graphical Processing Unit \\
        LYT & Lynetteholm Project \\
        RGB & Red, Green, Blue \\
        ROTV & Remotely Operated Towed Vehicles \\
        SQL & Structured Query Language \\
        TM & Temporal Mean \\ 
        URL & Uniform Resource Locator \\
        UW & Underwater \\
        ViT & Vision Transformer \\
        \hline
    \end{tabular}
    \end{adjustbox}
    \label{tab:glossary-acronyms}
\end{table}

\subsection{Screenshot of SeagrassFinder AP}

\begin{figure}[!ht]
    \centering
    \includegraphics[cfbox=black 1pt 1pt, width={0.6\columnwidth}]{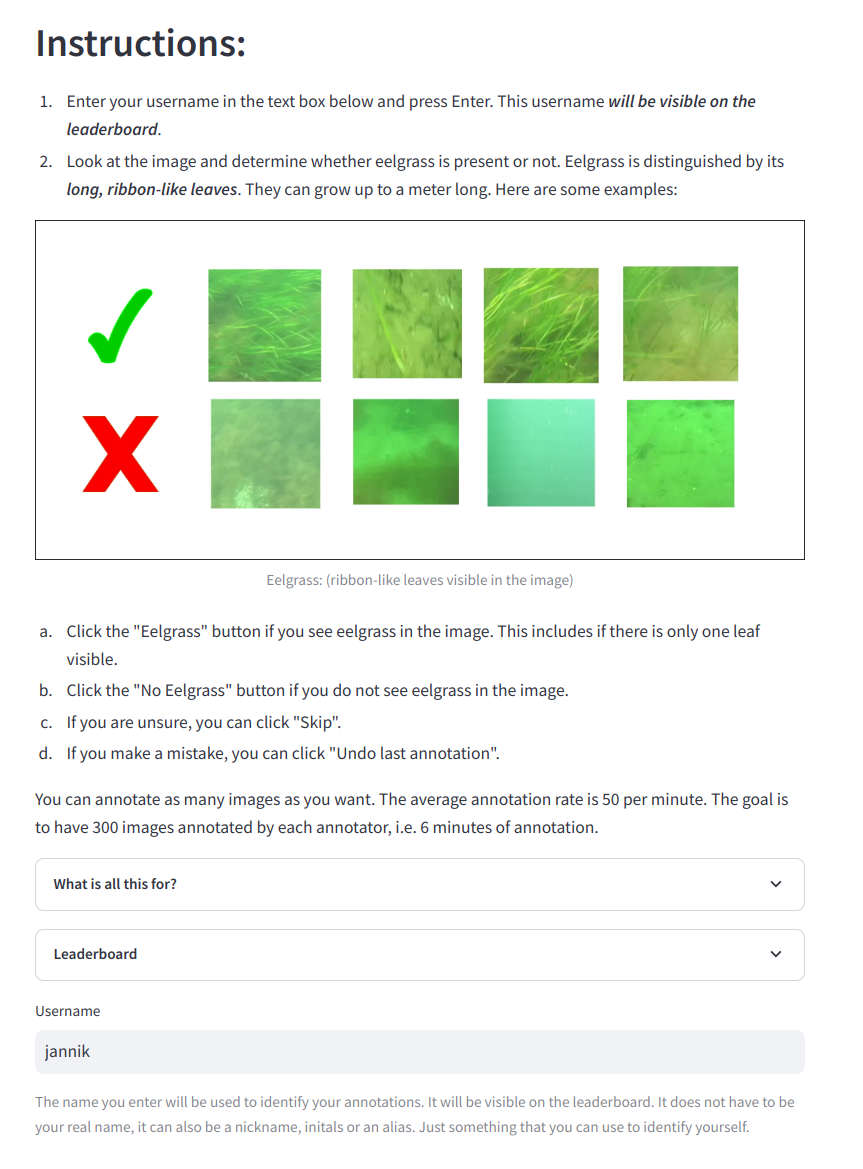}
    \caption{A screenshot of the start page of the SeagrassFinder AP showing the visual guidance for the labeling instructions.}
    \label{fig:seagrassfinder_visual}
\end{figure}

We provide a screenshot of the labeling instruction page of SeagrassFinder AP in \cref{fig:seagrassfinder_visual}.

\subsection{Eelgrass Presence by Water Depth}

\begin{figure}[ht]
    \centering
    \includegraphics[width=\ifbool{singlecolumn}{0.8\linewidth}{0.9\columnwidth}]{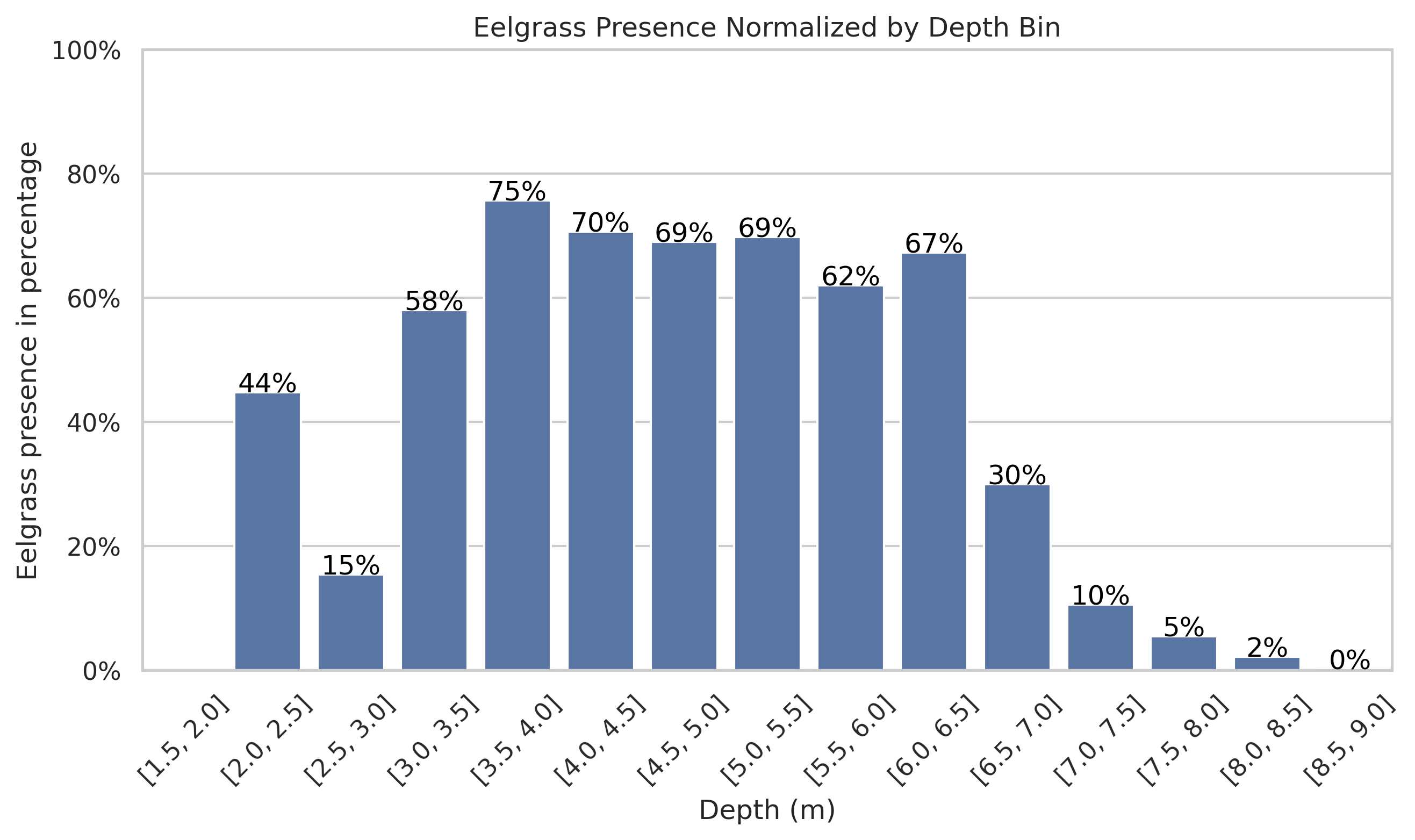}
    \caption{Eelgrass presence percentage sorted by depth (m).}
    \label{fig:lyn9-eelgrass-coverage}
\end{figure}

Based on the coverage estimations we can additionally extract eelgrass distribution relative to water depth, as it is an important factor in environmental impact assessments. This metric can be used as a proxy for estimating any human impact on the local marine ecosystem, as such impacts typically become most evident at the deepest limits of seagrass distribution. Transect surveys are specifically used to determine seagrass abundance over a range of water depths \citep{short2001global}. We present our results in \cref{fig:lyn9-eelgrass-coverage}.
\looseness -1

\bibliographystyle{plainnat}
\bibliography{bibliography}

\end{document}